\documentclass[sigconf]{acmart}

\copyrightyear{2024}
\acmYear{2024}
\setcopyright{rightsretained}
\acmConference[CIKM '24]{Proceedings of the 33rd ACM International Conference on Information and Knowledge Management}{October 21--25, 2024}{Boise, ID, USA}
\acmBooktitle{Proceedings of the 33rd ACM International Conference on Information and Knowledge Management (CIKM '24), October 21--25, 2024, Boise, ID, USA}
\acmDOI{10.1145/3627673.3679550}
\acmISBN{979-8-4007-0436-9/24/10}

\makeatletter
\gdef\@copyrightpermission{
  \begin{minipage}{0.3\columnwidth}
   \href{https://creativecommons.org/licenses/by/4.0/}{\includegraphics[width=0.90\textwidth]{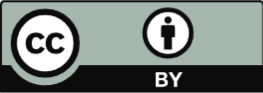}}
  \end{minipage}\hfill
  \begin{minipage}{0.7\columnwidth}
   \href{https://creativecommons.org/licenses/by/4.0/}{This work is licensed under a Creative Commons Attribution International 4.0 License.}
  \end{minipage}
  \vspace{5pt}
}
\makeatother

\usepackage{stfloats}
\usepackage{multirow}
\usepackage{arydshln}  
\usepackage{MnSymbol}
\usepackage{adjustbox}
\usepackage{booktabs}
\usepackage[ruled]{algorithm2e}
\usepackage{hyperref}
\usepackage{enumitem} 
\usepackage{amsmath}
\hypersetup{hidelinks,
	colorlinks=true,
	allcolors=black,
	pdfstartview=Fit,
	breaklinks=true}

\begin{document}

% !TEX TS-program = pdflatex

\title{Tackling Noisy Clients in Federated Learning with End-to-end Label Correction}


% \author{Xuefeng Jiang, Sheng Sun, Yuwei Wang, and Min Liu}
% \affiliation{%
%   \institution{
%     Institute of Computing Technology, Chinese Academy of Sciences, Beijing, China\\
%     University of Chinese Academy of Sciences, Beijing, China\\}
%   \city{}
%   \country{}
% }
% \email{{jiangxuefeng21b, sunsheng, ywwang, liumin}\textbf{@}ict.ac.cn}

% \author{Xuefeng Jiang}
% % \authornote{Both authors contributed equally to this research.}
% \email{jiangxuefeng21b\textbf{@}ict.ac.cn}
% \orcid{0000-0002-0211-9123}
% \author{Sheng Sun}
% \orcid{0000-0002-0260-2692}
% \affiliation{%
%   \institution{Institute of Computing Technology, Chinese Academy of Sciences \\ University of Chinese Academy of Sciences}
%   \city{Beijing}
%   \country{China}
% }

% \author{Yuwei Wang}
% \orcid{0000-0002-3228-7371}
% % \authornote{Both authors contributed equally to this research.}
% \author{Min Liu}
% \authornote{Corresponding author}
% \email{liumin\textbf{@}ict.ac.cn}
% \orcid{0000-0003-2824-9601}
% \affiliation{%
%   \institution{Institute of Computing Technology, Chinese Academy of Sciences \\ University of Chinese Academy of Sciences}
%   \city{Beijing}
%   \country{China}
% }

\author{Xuefeng Jiang}
\orcid{0000-0002-0211-9123}
% \author{Sheng Sun}
% \orcid{0000-0002-0260-2692}
\affiliation{%
  \institution{Institute of Computing Technology, Chinese Academy of Sciences \\ University of Chinese Academy of Sciences}
  \city{Beijing}
  \country{China}
}
\email{jiangxuefeng21b@ict.ac.cn}

\author{Sheng Sun}
\orcid{0000-0002-0260-2692}
% \author{Sheng Sun}
% \orcid{0000-0002-0260-2692}
\affiliation{%
  \institution{Institute of Computing Technology, Chinese Academy of Sciences}
  \city{Beijing}
  \country{China}
}
\email{sunsheng@ict.ac.cn}

\author{Jia Li}
\orcid{0009-0002-3987-5034}
% \orcid{0000-0002-0260-2692}
% \author{Sheng Sun}
% \orcid{0000-0002-0260-2692}
\affiliation{%
  \institution{Institute of Information Engineering, Chinese Academy of Sciences   \\ University of Chinese Academy of Sciences}
  \city{Beijing}
  \country{China}
}
\email{lijia1999@iie.ac.cn}

\author{Jingjing Xue}
\orcid{0000-0001-8328-3015}
\author{Runhan Li}
\orcid{0000-0002-8279-4750}
\author{Zhiyuan Wu}
\orcid{0000-0002-8925-4896}
% \orcid{0000-0002-0260-2692}
% \author{Sheng Sun}
% \orcid{0000-0002-0260-2692}
\affiliation{%
  \institution{Institute of Computing Technology, Chinese Academy of Sciences \\ University of Chinese Academy of Sciences}
  \city{Beijing}
  \country{China}
}
\email{xuejingjing20g@ict.ac.cn}

% \author{Zhiyuan Wu}
% % \author{Runhan Li}
% % \author{Zhiyuan Wu}
% % \orcid{0000-0002-0260-2692}
% % \author{Sheng Sun}
% % \orcid{0000-0002-0260-2692}
% \affiliation{%
%   \institution{Institute of Information Engineering, Chinese Academy of Sciences \\ University of Chinese Academy of Sciences}
%   \city{Beijing}
%   \country{China}
% }

\author{Gang Xu}
\orcid{0009-0002-3737-6477}
\author{Yuwei Wang}
\orcid{0000-0002-3228-7371}
% \author{Zhiyuan Wu}
% \orcid{0000-0002-0260-2692}
% \author{Sheng Sun}
% \orcid{0000-0002-0260-2692}
\affiliation{%
  \institution{Institute of Computing Technology, Chinese Academy of Sciences}
  \city{Beijing}
  \country{China}
}
\email{ywwang@ict.ac.cn}

% \author{Yuwei Wang}
% \orcid{0000-0002-3228-7371}
\author{Min Liu}
\authornote{Corresponding author}
\orcid{0000-0003-2824-9601}
\affiliation{%
  \institution{State Key Lab of Processors, Institute of Computing Technology, Chinese Academy of Sciences \\ University of Chinese Academy of Sciences \\ Zhongguancun Laboratory}
  \city{Beijing}
  \country{China}
}
\email{liumin@ict.ac.cn}

\renewcommand{\shortauthors}{Xuefeng Jiang et al.}

% \renewcommand{\shortauthors}{Xuefeng Jiang et al.}

\begin{abstract}
Recently, federated learning (FL) has achieved wide successes for diverse privacy-sensitive applications without sacrificing the sensitive private information of clients. However, the data quality of client datasets can not be guaranteed since corresponding annotations of different clients often contain complex label noise of varying degrees, which inevitably causes the performance degradation. Intuitively, the performance degradation is dominated by clients with higher noise rates since their trained models contain more misinformation from data, thus it is necessary to devise an effective optimization scheme to mitigate the negative impacts of these noisy clients. In this work, we propose a two-stage framework FedELC to tackle this complicated label noise issue. The first stage aims to guide the detection of noisy clients with higher label noise, while the second stage aims to correct the labels of noisy clients' data via an end-to-end label correction framework which is achieved by learning possible ground-truth labels of noisy clients' datasets via back propagation. We implement sixteen related methods and evaluate five datasets with three types of complicated label noise scenarios for a comprehensive comparison. Extensive experimental results demonstrate our proposed framework achieves superior performance than its counterparts for different scenarios. Additionally, we effectively improve the data quality of detected noisy clients' local datasets with our label correction framework. The code is available at \href{https://github.com/Sprinter1999/FedELC}{https://github.com/Sprinter1999/FedELC}.
\end{abstract}


% \ccsdesc[300]{Human-centered computing~Ubiquitous and mobile computing}




\maketitle

\section{Introduction}

The pervasiveness of mobile devices contributes more than half of the internet traffic \cite{flexifed}, which empowers a variety of intelligent applications.
To exploit these distributed datasets from these mobile clients in a privacy-preserving manner, federated learning (FL) \cite{fedSummary,fedavg,logitFusion} has evolved as a promising collaborative training paradigm, which has shown significant successes in real-world applications like health-care \cite{fedeye,FedNoRo} and recommendation system \cite{recomm,tanben}.

Data are often not independent identically distributed (Non-IID) across all clients, and many methods have been proposed to  address the data heterogeneity or challenge in recent years. 
Another potential issue concerning incorrect labels (a.k.a. noisy labels) contained in client-side datasets has been long overlooked.
Specifically, it is hard to promise the client-side datasets are well-annotated in practice.
%However, another potential problem regarding clients' data in FL has been long neglected. 
%Specifically, it is hard to promise the client-side local datasets are well-annotated in practice. 
%In other words, client-side datasets inevitably contain incorrect labels (a.k.a. noisy labels).
Each client is usually not given enough incentives \cite{incentives} to provide a well-annotated dataset, while the server is not authorized to directly and accurately inspect the data quality of the client datasets due to privacy concerns \cite{FedLSR}. 
More commonly, these client-side datasets tend to contain varying degrees of label noise since different clients have diverse approaches to collecting annotations or have respective limited budgets to annotate their datasets.
As deep neural networks have the capacity to fit local datasets containing noisy labels \cite{memorization}, the negative information from incorrect data can be injected into model parameters, further perturbing the overall training of FL.
%As deep neural networks have the high capacity to fit noisy labels \cite{overfit,co-teaching}, it is not difficult for these networks to fit the local datasets containing noisy labels, which injects the negative information from the data into the model parameters and further perturb the overall training of FL.
Therefore, data with noisy labels pose critical challenges in the real-world applications of FL, which evidently decrease the convergence rate and final performance of FL \cite{flr}. 
To tackle these challenges, there have been some early attempts \cite{FedLSR,rhfl,robustfl} to improve the trained model's robustness against noisy labels in the recent two years. However, these methods often neglect one challenge that different clients tend to have distinctly inconsistent label noise rates. Specifically, different clients tend to have different annotation quality in practice \cite{flr,fedcorr}. Even worse, some underlying malicious clients might aim to provide low-quality datasets containing high label noise rate to perturb the FL training on purpose \cite{robustfed}. 
%Therefore, a more practical setting is to simultaneously suppose the client-side datasets are non-IID and of different noise rates. 
In the practical noisy scenario, clients with high noise rates play detrimental roles in the model performance of FL since more negative information is injected into their trained models caused by data with higher label noise \cite{flr}. 
Therefore, it is of significance to devise the specific training scheme for noisy clients with higher noise rates.
One recent state-of-the-art method FedNoRo \cite{FedNoRo} attempts to detect noisy clients via exploiting a two-component Gaussian mixture model \cite{gmm} based on fine-grained indicators, and then applies different robust loss terms for detected clean clients and noisy clients.
%To deal with heterogeneous and noisy clients, one recent state-of-the-art method FedNoRo \cite{FedNoRo} provides the insight of detecting noisy clients via exploiting a two-component Gaussian mixture model \cite{gmm} based on fine-grained indicators, and then utilizes robust loss terms to prevent the negative effects caused by noisy clients.
%proposes to calculate per-class average loss values as indicators and exploit a two-component Gaussian mixture model \cite{gmm} based on these indicators to detect noisy clients 
%which empirically shows a high detection accuracy of noisy clients. 
%After identifying the noisy clients, FedNoRo designs robust loss terms based on knowledge distillation \cite{KD} specifically for these noisy clients to prevent the negative effects caused by noisy clients. 
All aforementioned methods only focus on improving robustness against the label noise, however, they neglect the potential of correcting the underlying noisy samples in client-side datasets to further improve model performance of FL, especially for clients with higher label noise rates.
%they neglect to explore the potential of correcting the underlying noisy samples in client-side datasets, especially for clients with higher label noise rates.

%However, most existing methods mainly aim to improve robustness against the label noise but lack an explicit approach to effectively correcting the underlying noisy samples of client-side datasets, especially for clients with higher label noise rates.
In this work, we focus on the challenge of FL regarding data quality, where the clients can have heterogeneous data and different noise rates. 
%We explore if there is a possibility to simultaneously improve the data quality of clients with higher label noise, which can also further improve the robustness of FL model. 
%We refer to FedNoRo to divide the clients into the relatively clean group and the relatively noisy group. 
The participating clients can be divided into the relatively clean group and the relatively noisy group, referring to FedNoRo \cite{FedNoRo}.
Considering that the detected noisy clients have more negative impacts on the final performance of the trained model, we exploit an end-to-end label correction scheme to gradually refine the corrupted local datasets of the detected noisy clients. 
Specifically, a differentiable variable is designed to model the possible underlying ground-truth label distribution of each sample.
%which is initialized from the original one-hot label annotation. 
We incorporate this variable into the optimization objectives of the local update procedure, and both the model and this variable can be simultaneously updated through the back propagation.
%We introduce local estimated label distribution as a differentiable variable for each sample, which is initialized from the original one-hot label and updated by the back propagation.
%  , aiming to locally obtain extra supervision from data.
%which benefits from data augmentation technique to mitigate the negative effect caused by noisy labels.
%Specifically, we implicitly restrain local training with extra supervised information by mixing the sharpened prediction of original and augmented samples.
%In addition, we further utilize self knowledge distillation technique to explicitly minimize the prediction discrepancy between original and augmented samples.
%The underlying motivation is that we can obtain extra supervision information from the augmented instance.
To put it briefly, our main contributions can be summarized as follows:
\begin{itemize}[leftmargin=0.3cm]
% \begin{itemize}
  % \item Considering the inevitability of noisy labels in federated learning, we suppose to simultaneously improve the robustness of the trained model and also correct the noisy labels.
  \item We present a two-stage robust FL framework to tackle clients of different noise rates.
  The first stage aims to detect the underlying noisy clients and the second stage aims to correct labels of detected noisy clients via end-to-end label optimization.  
  \item We implement sixteen baseline methods for comparison, and showcase the superiority of our proposed method against its counterparts through extensive experiments on multiple label noise scenarios derived from four benchmark datasets.
  \item We further evaluate the effectiveness of our method on one large-scale real-world Clothing1M dataset, which contains one million images crawled from several famous shopping websites annotated with complicated systematic noisy labels.
\end{itemize}

% The rest of this article has been organized in the following way: Section \ref{RW} discusses the related works. Section \ref{sec:method} presents problem definition and methodology of our approach, and section \ref{sec:exp} shows the empirical evaluation and in-depth analysis on four benchmark datasets and one real-world dataset. Section \ref{conclude} concludes this article.

\section{Related Works}
\label{RW}
\subsection{Federated Learning}
Federated learning (FL) is an emerging distributed machine learning paradigm across clients over the network, which provides an opportunity to further exploit the distributed client datasets \cite{junxu,fedavg,fedrn}. 
One general issue in such distributed datasets is data distributions from different clients are usually non-identically distributed (non-IID). 
There are many previous works addressing this issue and herein we discuss some representative methods. 
FedAvg \cite{fedavg} averages updated model parameters to aggregate diverse information from client datasets. 
FedProx \cite{fedprox} proposes a proximal term during the client-side local update phase to restrain the local model parameters from deviating too much from the received global model parameters of the current communication round. Favor \cite{wanghao} aims to learn an optimal client subset selection strategy based on deep Q-learning \cite{qlearning}. FedMC \cite{fedmc} proposes a personalized FL optimization framework via meta learning \cite{metalearning} which also supports heterogeneous local model architectures. More recently, FedExP \cite{FedExP} speeds up FedAvg via parameter extrapolation on the server side. To support training with lower computation costs in the deployment of FL, some works propose light-weight training strategies like model pruning and parameter sparsification \cite{biad,skel,fedpse}. 
Above methods mainly focus on the non-IID issue of distributed client datasets, which we name by \underline{general FL methods}. However, besides the non-IID issue, another challenge is long neglected. That is, to provide a high-quality annotated dataset is too expensive \cite{NLL} which requires much human effort and expertise. Some methods like utilizing crowd-sourcing  \cite{robustfed} and machine-generated labels \cite{machine-gene, robustfl,FedLSR} seem to be lower-cost approaches to annotating a collected dataset. In the specific context of FL,  since the central server cannot enforce clients' behavior \cite{robustfed}, data are more easily to be annotated with noisy labels.  Meanwhile, owing to the extremely high expressive power of deep networks, networks can memorize training data even when labels are extremely noisy \cite{selfie}.  Training on data with wrong labels (i.e. noisy labels) evidently decreases the convergence rate and final performance \cite{FedLSR,robustfl}.

\subsection{Noisy Label Learning}
To tackle data with noisy labels is widely discussed in centralized learning scenarios, so we firstly discuss some representative \underline{noisy label learning (NLL) methods}  \cite{NLL,NLLnara,hbsur}.
One prominent research direction aims to conduct sample selection. Co-teaching \cite{co-teaching} and its derivative method Co-teaching+ \cite{co-teaching+} simultaneously maintain two peer networks of the same architecture with different initialization. For each batch, each network selects samples from the batch to its peer network according to different techniques. For instance, Co-teaching supposes samples with lower losses are more reliable and they are selected and fed into its peer network to perform training with more reliable supervision information.  

Another prominent research line focuses on designing robust loss \cite{symmetricCE} or robust training strategies \cite{jointopt,selfie,Dividemix}. 
Since the given labels can be noisy and the model can give correct predictions, Symmetric Cross Entropy (Symmetric CE \cite{symmetricCE}) includes the model prediction into loss terms. 
Tanaka et al. \cite{jointopt} propose joint optimization framework (Joint Optim), which can correct labels during training by alternating update of network parameters and labels. 
The labels are gradually updated by averaging the predictions made by the trained model in previous epochs.  
SELFIE \cite{selfie} robustly selects potential unclean samples and gradually add them into the training process. DivideMix \cite{Dividemix} integrates multiple techniques including Co-teaching, data augmentation method MixUp \cite{mixup} and one semi-supervised learning \cite{fedloke} framework MixMatch \cite{mixmatch}. 

Most of the above mentioned NLL methods can be simply incorporated into the FL pipeline since we can combine them with the most ubiquitous model aggregation method FedAvg \cite{fedavg}, which are often included as baseline methods in previous FL works \cite{FedLSR,robustfl,fedrn}. In this work, inspired by Joint Optim \cite{jointopt}, we try to effectively correct the labels of detected noisy clients' dataset since the noisy clients tend to have worse impacts on the FL training.

\subsection{Federated Noisy Label Learning}
\label{sec:fnll}

In the community of FL, to address the label noise issue, a simple research line of studies aims to design \underline{robust aggregation methods} \cite{krum,median,TrimmedMean} instead of simply averaging the model parameters used in many previous works \cite{fedavg,fedprox}. One commonly utilized aggregation method  Median \cite{median}  is based on the median value of the clients' models, instead of the weighted averaged value as FedAvg. In this way, extremely bad weights may affect less to the global model. Similarly, TrimmedMean \cite{TrimmedMean} removes the largest and smallest one for each parameter of selected clients' models, and computes the mean of the remaining parameters as the global model. Krum \cite{krum} firstly computes the nearest neighbors to each local model. Then, it calculates the sum of the distance between each client and their closest local models. Finally, it selects the local model with the smallest sum of distance as the global model. 
These methods mainly focus on decreasing the effects of bad model weights caused by noisy labels via more robust aggregation methods, but they do not directly tackle the data with noisy labels.
There are also some emerging methods that aim to directly deal with the noisy labels in the context of FL, and we name them by \underline{federated noisy label learning (FNLL) methods} \cite{FedNoRo,FedLSR}. To the best of our knowledge, Robust FL \cite{robustfl} is the first work to directly deal with data with noisy labels without the requirement of extra perfectly-annotated auxiliary dataset \cite{FedLSR,fedcorr}, which collects the local class-wise centroids to form global averaged class-wise centroids as extra global supervision to regularize the local training phase. However, the transmitted class-wise centroids carry sensitive information of client data, which can be easily exploited to reveal private information via inverse engineering \cite{decorr,FedLSR}. To further maintain data privacy, FedLSR \cite{FedLSR} proposes a local regularization method based on self-distillation \cite{allenZhu}. FedRN \cite{fedrn} maintains a server-side client model pool to exploit reliable neighbor models (RN) for each client, where the RNs refer to the models owned by other clients that possess data of similar distributions or lower label noise. FedRN then detects clean samples for each client, using ensembled Gaussian Mixture Models trained to fit loss function values of local data assessed by these RNs. A recent work FedNoRo \cite{FedNoRo} proposes a two-stage framework to identify the noisy clients and design different local optimization objectives for clean clients and noisy clients, which reports a satisfying noisy client detection accuracy and final performance of the trained model. Different from the above FNLL methods, RHFL \cite{rhfl} supports different client model architectures with the assistance of an auxiliary dataset which is a hard assumption in practice, so we do not include RHFL for evaluations.

Meanwhile, we find the experimental settings are not unified to a certain degree. For example, some works \cite{FedLSR,rhfl,robustfl} neglect the Non-IID characteristic of FL. Therefore, we cover more general settings that approximate the diverse label noise scenarios in practice. We prepare Table \ref{tab:comparison_prev} to better illustrate the differences (especially in the methodology and experimental settings) between our method and related works that we implement and evaluate in this study.
\begin{table}[tbp]
\caption{Comparison of baselines regarding  properties of methodology (M) and experimental settings (E) : (M1) sample or client selection (M2) robust aggregation, (M3) robust loss, (M4) explicit label correction,  (E1) data heterogeneity, (E2) label noise, (E3) label noise of varying levels across clients and (E4) mixed label noise patterns (see Section \ref{sec:labelnoise}).}
\label{tab:comparison_prev}
\begin{adjustbox}{width=\columnwidth,center}
\begin{tabular}{c|cccc|cccc}
\hline 
Methods & M1 & M2 & M3 & M4 & E1 & E2 & E3 & E4 \\ \hline
FedAvg \cite{fedavg} &  &  &  &  & $\checkmark$ &  & & \\
FedProx \cite{fedprox}&  &  &  &  & $\checkmark$ &  & & \\
FedExP \cite{FedExP}&  &  &  &  & $\checkmark$ &  &  & \\
TrimmedMean \cite{TrimmedMean} &  & $\checkmark$ &  &  &  & $\checkmark$ & & \\
Krum \cite{krum} &  & $\checkmark$ &  &  &  & $\checkmark$ &  & \\
Median \cite{median} &  & $\checkmark$ &  &  &  & $\checkmark$ & & \\
Co-teaching \cite{co-teaching} & $\checkmark$ &  & $\checkmark$ &  &  & $\checkmark$ & & \\
Co-teaching+ \cite{co-teaching+}& $\checkmark$ &  & $\checkmark$ &  &  & $\checkmark$ & & \\
Joint Optim \cite{jointopt}&  &  & $\checkmark$ & $\checkmark$ &  & $\checkmark$ &  & \\
SELFIE \cite{selfie}& $\checkmark$ &  & $\checkmark$ &  &  & $\checkmark$ & & \\
Symmetric CE \cite{symmetricCE}&  &  & $\checkmark$ &  &  & $\checkmark$ &  & \\
DivideMix \cite{Dividemix} & $\checkmark$ &  & $\checkmark$ &  &  & $\checkmark$ & & \\
Robust FL \cite{robustfl}& $\checkmark$ &  & $\checkmark$ &  &  & $\checkmark$ & $\checkmark$ & \\
FedLSR \cite{FedLSR}&  &  & $\checkmark$ &  &  & $\checkmark$ &  & \\
FedRN \cite{fedrn}&  &  &  &  & $\checkmark$ & $\checkmark$ & $\checkmark$ &$\checkmark$ \\
FedNoRo \cite{FedNoRo}&  $\checkmark$ & $\checkmark$ & $\checkmark$ &  & $\checkmark$ & $\checkmark$ & $\checkmark$ &\\ \hline
Ours (FedELC) & $\checkmark$ & $\checkmark$ & $\checkmark$ & $\checkmark$ & $\checkmark$ & $\checkmark$ & $\checkmark$ & $\checkmark$ \\ \hline
\end{tabular}
\end{adjustbox}
\end{table}

% Table 4: Comparison of previous FNL methods according to several properties: (P1) sample selection (P2) robust aggregation, (P3) label correction, (P4) robust loss, (P5) no supervision, (P6) data heterogeneity, and (P7) heterogeneous noise levels.

%can be inversely utilized to reveal important private information which can cause serious privacy issues. 

\section{Preliminary}

\subsection{Problem Definition}
Without loss of generality, assume the federated learning (FL) system consists of $N$ clients and a server. 
Let $\mathcal{S}$ represent the set for all $N$ clients.
Each client indexed by $k$ maintains a local dataset with $n_k$ samples $\mathcal{D}_{k}=\left\{\left(x_{i}, \hat{y}_{i}\right)\right\}_{i=1}^{n_{k}}$, and we denote the total data quantity as $n=\sum_{k\in S}n_k$. Note that the one-hot label $\hat{y}_{i}$ can be noisy and we denote the unknown real ground-truth label is $y^*$. The overall objective of FL is solving the optimization problem for $N$ clients over their own local datasets, which can be formulated as:
\begin{equation}
    \min_{\boldsymbol{w}}f(\boldsymbol{w}):=\sum_{k\in S}\frac{n_k}nF_k(\boldsymbol{w}).
\end{equation}
Local objective of client $k$ is to minimize empirical risks on $D_k$:
\begin{equation}
    F_k(w)=\mathbb{E}_{(x,\hat{y})\sim\mathcal{D}_k}\left[{\ell}_k(\hat{y},f_k(x;w_k))\right],
\end{equation}
where $\ell_k$ is the loss function on the $k-th$ client local dataset and $f_{k}(x;w_k)$ is the local prediction on each sample $x$ with the local trained model parameterized by  $w_k$. 

It is assumed that $k-th$ client has its own Non-IID dataset $\mathcal{D}_{\mathrm{k}}$, and cannot directly access another client's data. Following the general FL framework \cite{fedavg}, FL requires multiple global rounds $T$ to achieve the final convergence. In each global round $t$, the server firstly selects a group of clients $\mathcal{S}_t$ to perform local update  for $E$ local epochs, and then it aggregates the updated local models from $\mathcal{S}_t$ to form the updated global model $w^{t+1}$:
\begin{equation}
\label{eq:fedavg}
    w^{t+1}=\sum_{k\in S_{t}}\frac{n_{k}}{n_t}w_{k}^{t},
\end{equation}
where $n_t$ indicates the total number of data of all selected clients $\mathcal{S}_t$  selected in the global round $t$.

% \subsection{Data Heterogeneity}
% In order to consider the data heterogeneity in FL, we first consider the data distribution across all clients prior to introducing label noise which is a common setting among the related works \cite{chaos,flr}. Following the common practice in previous works \cite{fedrn,FedExP,decorr}, we utilize Dirichlet distribution to distribute the total data to different clients: each client is assigned with training data drawn from the Dirichlet distribution with a concentration parameter $\beta$, where a lower $\beta$ indicates a higher Non-IID degree, which introduces higher difficulties for FL training \cite{qinbinCrossSilo}.

\subsection{Label Noise Patterns}
\label{sec:labelnoise}
Herein we discuss three common label noise patterns in our study: 

\textbf{Manually-injected label noise}
We use the label transition matrix $\mathcal{M}$ to add manual label noise to local datasets, where $\mathcal{M}_{i,j}=flip(\hat{y}=j|y^*=i)$ represents that the true ground-truth label $y^*$ is flipped from the clean class to the noisy class $\hat{y}$.
There are two commonly-used structures for the matrix $\mathcal{M}$ \cite{FedLSR,rhfl,robustfl}, symmetric flipping and asymmetric (or pairwise) flipping \cite{making,co-teaching}. Symmetric flipping means that the original class label will be flipped to any wrong class labels with equal probability. 
For asymmetric flipping, it means that the original class label will only be flipped to another wrong category. 
In FL, given the maximum noise rate $\epsilon$, we generate each client $k$ with corresponding noise level increasing linearly from $0$ to $\epsilon$ as the client's index $k$ increases.
Besides the symmetric and asymmetric label noise scenarios, inspired by FedRN \cite{fedrn}, we also generate mixed label noise scenario where clients are divided into two halves, and one half of clients follow the symmetric label noise and the remainder follows the asymmetric label noise. 

\textbf{Human annotation error caused label noise}
Evaluations on datasets approximating real-world human annotation error patterns are less considered in previous works. Fortunately, Amazon Mechanical Turk provides CIFAR-N \cite{cifarn} datasets, which collect labels purely from human annotators. These datasets begin to be utilized in some recent attempts  \cite{chaos,flr,cifarn}. There are several versions of CIFAR-N datasets, and we utilize the CIFAR-10-N-Worst and CIFAR-100-N-Fine since they possess the highest noise rates. The noise rates are 40.208\% and 40.200\%, respectively.

\textbf{Systematic label noise} In practice, it is expensive to obtain high-quality labels for large-scale datasets. One commonly-utilized method is to collect images from websites and yield annotations by filtering out the surrounding context (e.g. image captions \cite{clip}) in the web page. Clothing1M \cite{clothing1m} is a large-scale real-world dataset of 14 categories, which contains 1 million images of clothing with noisy labels crawled from several famous online shopping websites. It is reported that the overall noise ratio is approximately 38.46\% and contains unstructured complicated label noise \cite{augdesc}.

\section{Method}
\label{sec:method}
To deal with the complicated issue introduced by heterogeneous and noisy clients, we propose a two-stage method FedELC as illustrated in Figure \ref{fig:elc}. During the first stage, all clients are orchestrated to follow the FedAvg \cite{fedavg} procedure to aggregate client updated models, which is used in previous works \cite{FedNoRo,FedLSR,robustfl}. The first stage (stage \#1) mainly aims to divide the total clients into one relatively clean group and the other relatively noisy group. The second stage (stage \#2) mainly aims to gradually correct the data with noisy labels located at noisy clients which are detected in the stage\#1. 

\begin{figure}[bp]
    \centering
    \includegraphics[width=\linewidth]{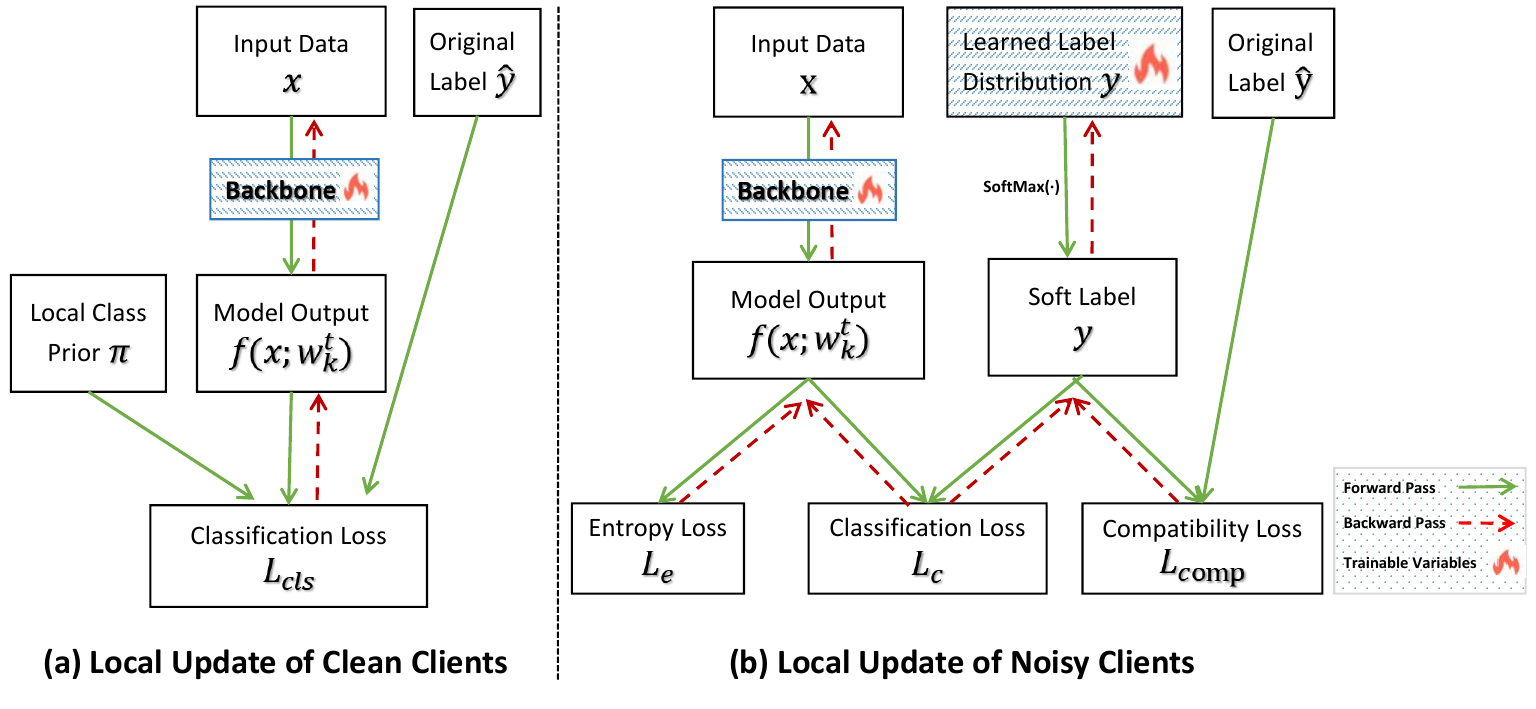}
    \caption{Illustration of local update procedures.}
    \label{fig:elc}
\end{figure}

\subsection{Stage \#1: Noisy Client Detection}
Our method is divided into two stages where stage \#1 lasts for $T_w$ global communication rounds. We follow FedNoRo \cite{FedNoRo} to perform the first stage. During the stage\#1,  a warm-up model is firstly trained based on FedAvg \cite{fedavg}. When the stage\#1 ends, we exploit a two-component Gaussian Mixture Model (GMM) to divide total clients into a relatively clean group and a relatively noisy group.

For the each round $t$ in the first stage, each selected client $k$ updates the local model with the vanilla cross entropy loss and conduct model aggregation with FedAvg in Eq. \ref{eq:fedavg}. Different from the original FedAvg \cite{fedavg}, considering the local data heterogeneity, each selected client firstly computes the local class distribution $\pi$ of the local dataset, which is often used as class prior. We use this class prior $\pi$ to conduct logit adjustment \cite{LA} before we compute the vanilla cross entropy loss, which is demonstrated to be effective in previous works \cite{tao2023local,adaptiveLA,LA} and used in FedNoRo. As discussed in \cite{FedNoRo}, logit adjustment is to make the local model treat each class equally rather than being biased. We use $\theta$ to denote the current trained local model weight of $w_t^k$. For a sample $(x,\hat{y})$, the model yields the output prediction $p=f(x;\theta)$. We use this prior $\pi$ to adjust the logit into $p + log(\pi)$ before computing cross entropy (CE) loss, and then we formulate the local training objective as:

% \vspace{2pt}
\begin{equation}
\label{eq:cls}
    \mathcal{L}_{cls} = CE(p,\pi,\hat{y})
\end{equation}

After $T_w$ global training rounds, we conduct the noisy client detection. Many previous works aim to conduct noisy client detection \cite{chen2020focus} with the overall local loss value. For example, one commonly used assumption is that samples with higher loss are more likely to be mistakenly labeled \cite{co-teaching,robustfl,memorization}, thus clients with higher loss are more possible to be noisy clients. However, since each client in FL possesses Non-IID data, such coarse grained division technique can be less effective as analyzed in \cite{FedNoRo,fedcorr}.

Following FedNoRo \cite{FedNoRo}, we consider one fine-grained class-aware noisy client detection method. For each client indexed by $k$ participating in FL training, each client used the global model $w^{T_w}$ of $T_w-th$ global round to compute the local class-wise loss. The average loss values of all classes on each client $k$ denoted as $\ell_k = {(\ell_k^1, \ell_k^2,\ldots, \ell_k^M)} \in \mathbb{R}^M $ do not contain private information, and they are transmitted to the server for further noisy client detection. Since we have $N$ clients in total, now we construct a loss matrix $L=[\ell_1,\ell_2,\ell_3,\ldots,\ell_N] \in \mathbb{R}^{N*M}$, which reflects the client-wise per-class loss vectors. Then, a two-component GMM is deployed onto $L$ to partition all $N$ clients into two subsets: the relatively clean group $S_{clean}$ and detected relatively noisy group $S_{noisy}$. More details are available in \cite{FedNoRo}. We analyze its performance in Section \ref{sec:detection}.

\subsection{Stage \#2: End-to-end Label Correction} 
\label{sec:stage2}
\begin{algorithm}[tbp] 
    \caption{Local Update Procedure of FedELC}
    \label{algo}
    \LinesNumbered
    \KwIn{client $k$ , current global round $t$, global model $w_t$ }
    \KwOut{local trained model $w_t^k$}%输出
    $w_t^k \gets w_t$  \; 
    \If{t < $T_w$ or $k$ is clean client}
    {
     \textcolor{blue}{/* Vanilla training*/}\\
     Compute local class distribution prior $\pi$ \; 
    \For{each local epoch $i$ from 1 to $E$}{
        \For{each batch $(x,\hat{y})$ }{
        % $o_1, o_2 = f(x;w_t^k), f(Augment(x);w_t^k)$\;
        % $o_2 = f(Augment(x);w_t^k)$\;
        % $p_1, p_2 = SoftMax(o_1), SoftMax(o_2)$ \; 
        % $\lambda \sim Beta(1,1)$ \;
        % $p = \lambda*p_1 + (1-\lambda)*p_2$  \;
        % $p_{s} = Sharpen(p,T)$ \;
        % \textcolor{blue}{/* Phase 2: FedAvg with FLR loss*/}\\
        $p = f(x; w_t^k)$           \; 
        $\mathcal{L}_{cls}$ = CE($p$, $\hat{y}$, $\pi$) ; \textcolor{blue}{// Eq. \ref{eq:cls}}  \\
        % $Loss_{reg} = SelfDistillation(o_1,o_2, T_d)$ \;
        % $Loss = Loss_{cls} + \gamma * Loss_{reg} $ \;
        Update $w_t^k$ with $\mathcal{L}_{cls}$ via \textit{Back Propagation} \;
        }
    }
    }
    \Else{  
    \textcolor{blue}{/* End-to-end label correction for noisy clients*/}\\
    Initialize $y_d$ with original label $\hat{y}$ \;
    \For{each local epoch $i$ from 1 to $E$}{
        \For{each batch $(x,y)$ }{
        $p = f(x;w_t^k)$\;
        % \textcolor{blue}{/* Phase 2: FedAvg with FLR loss*/}\\
        $\mathcal{L}_{c} = CE(p,y^d)$ ;  \textcolor{blue}{// Eq. \ref{eq:classification}}  \\ 
        $\mathcal{L}_{comp} = Compatibility(\hat{y},\tilde{y})$  ; \textcolor{blue}{// Eq. \ref{eq:comp}}  \\
        $\mathcal{L}_{e} = Entropy(p)$ ; \textcolor{blue}{// Eq. \ref{eq:entropy}}  \\
        $\mathcal{L} = \mathcal{L}_{c} + \alpha * \mathcal{L}_{comp} + \beta * \mathcal{L}_{e} $ ; \textcolor{blue}{// Eq. \ref{eq:all_loss}}  \\
        Update $w_t^k$ and $\tilde{y}$ with $\mathcal{L}$ via \textit{Back Propagation} \;
        % Update $$ with $Loss$ via \textit{Back Propagation} \;
        }
    }
        \textcolor{blue}{/*Merge learned labels and updated model's prediction*/} \\
    \For{each local sample x}{
    $P_{model} = SoftMax(f(x; w_t^k))$  \;
    % $P_{predict} = SoftMax(P_{logit}) $\;
    $\tilde{y} = \frac{P_{model} + SoftMax(\tilde{y})}{2} \times K$ ; \textcolor{blue}{// $y^{d}$  is also updated} \\
    $y_{estimate} = SoftMax(\tilde{y})$ ; \textcolor{blue}{// Estimate possible $y^{*}$}  \\
    }
    }
\end{algorithm}

Now we have two groups $S_{clean}$ and $S_{noisy}$. For clients in $S_{clean}$, we use the same optimization objective in Eq. \ref{eq:cls} to conduct local update (more explanations in Section \ref{sec:cifar-m}). For clients in $S_{noisy}$, to further exploit data from detected noisy clients, we adopt the following method to conduct both robust local model update and local data refinery by end-to-end correcting the local labels inspired by \cite{pencil} which shows promising noisy label refining performance. 

\textbf{End-to-end label correction}
For starters, we clarify concepts of hard/soft label for M-way classification. We denote $\mathbf{1}$ is a vector of all-ones. For the label of a certain sample, the one-hot hard label space is $\mathcal{H}=\{y:y\in\{0,1\}^m,\mathbf{1}^\top y=1\}$, and the soft label space is $\mathcal{S}=\{y:y\in[0,1]^m,\mathbf{1}^\top y=1\}$, which reflects a probabilistic label distribution. 
For a sample $(x,\hat{y})$ in the local dataset, we suppose it has the unknown ground-truth label $y^*$, and both $y^*$ and $\hat{y}$ is a one-hot hard label. We initialize a distribution $y^d \in \mathcal{S} = \{y:y\in[0,1]^m,1^\top y=1\}$ to denote our estimation of the underlying pseudo-ground-truth soft label distribution, which is initialized by the original $\hat{y}$ and can be gradually updated via back propagation:

\begin{equation}
    \tilde{y}=K\hat{y},
\end{equation}

\begin{equation}
    y^d=\operatorname{SoftMax}(\tilde{y}),
\end{equation}
\
where $\tilde{y}$ is a differentiable variable which can be updated by back propagation and K is a large constant (K = 10 by directly following the original implementation of \cite{pencil}).

For noisy clients, the local optimization objective is combined of three terms.
% There are also some methods like Joint Optim \cite{jointopt} considers directly corrects the noisy labels  via exploiting the model prediction as the updated labels. However, through our wide evaluation on diverse types of label noise scenarios (see Table \ref{tab:results_c10}, Table \ref{tab:results_c100}, Table \ref{tab:cifar10-n}, and Table \ref{tab:cifar100-n}), we find this vanilla label correction can be less effective in FL since clients share Non-IID data, and it can be hard to local models to give correct predictions because the updated local models can be biased due to data heterogeneity. 
% yd i is used as the pseudo-groundtruth label in our learning, which is initialized based on the noisy label ˆyi. It is continuously updated (i.e., the noise is gradually corrected) through backpropagation. This probabilistic setting allows ample flexibility for noise correction.
The first one is the basic classification loss $\mathcal{L}_{c}$. Rather than using the original label $\hat{y}$, we compute the classification loss between the model prediction $p$ and the learnable distribution $\tilde{y}$:

\begin{equation}
\label{eq:classification}
    \mathcal{L}_{c} = CE(p,\tilde{y})
\end{equation}

The second one is compatibility regularization loss. For most scenarios where the noise levels of most clients are not too high, the original label $\hat{y}$ can also provide useful supervision \cite{pencil}. In other words, for the overall samples, the optimized distribution $\tilde{y}$ should not be very far from the original labels $\hat{y}$. Therefore, the compatibility regularization loss can be formulated as:

\begin{equation}
\label{eq:comp}
    \mathcal{L}_{comp} = Compatibility(\hat{y},\tilde{y}) = -\sum_{m=1}^M\hat{y}_{m}\log (y_{m}^d).
\end{equation}

The third one is entropy regularization loss, which is used in FedLSR \cite{FedLSR}, Robust FL \cite{robustfl} and semi-supervised learning methods \cite{Dividemix,entropyy}. The main target is to encourage the model to output sharper and more confident prediction, which can be formulated as:
\begin{equation}
\label{eq:entropy}
     \mathcal{L}_e= Entropy(p) =-\sum_{m=1}^Mp^{m}\log(p^{m}),
\end{equation}
% \vspace{-1pt}
where $p_{model}^{m}$ is the model’s output softmax probability for the $m-th$ class.
Combining all three terms together, we form triplet supervision for the detected noisy clients as follows

\begin{equation}
    \label{eq:all_loss}
    \mathcal{L} = \mathcal{L}_{c} + \alpha * \mathcal{L}_{comp} + \beta * \mathcal{L}_{e},
\end{equation}
where $\alpha$ and $\beta$ represent the trade-off coefficients to balance these loss terms. $\mathcal{L}_{c}$ and $\mathcal{L}_{comp}$ provide supervisions to optimze the learnable $\tilde{y}$, thus, the optimzation process of $\tilde{y}$ can be formulated as

\begin{equation}
\label{eq:backward_y}
    \tilde{y} = \tilde{y} - \eta*\nabla\tilde{y}
\end{equation}

Note that the learning rate $\eta$ for $\tilde{y}$ is different from the learning rate for the optimized model $\theta$. We elaborate more on its selection in Section \ref{sec:ablation}. After $E$ epochs of local updating, we can provide our estimation of the corrected label which is an approximation of the possible ground-truth label $y^*$ via fusing two ways of estimation including the trained model's prediction and the updated $\tilde{y}$.

We record the wall-clock training time of the clean clients and noisy clients, and on average the training time of noisy clients is almost $\times0.3$ longer than clean clients. Therefore, considering the computation cost and noisy clients can have higher risks to bring negative impacts to FL training, we choose to apply the end-to-end correction procedures only to the detected noisy clients.

\textbf{Distance-aware (DA) aggregation}
In the second stage, besides the label correction operation, we refer to exploit distance-aware aggregation method instead of the vanilla model averaging of FedAvg \cite{fedavg}.
Considering the co-existing of the clean and the noisy clients, intuitively, models from clean and noisy clients are of different importance.
Inspired by robust aggregation method Krum \cite{krum} and previous FL method \cite{FedNoRo, rscfed}, we utilize one commonly-used distance-aware client-wise distance metric is defined as
\begin{equation}
    d(i)=\min_{j\in S_t}\|w_i^t-w_j^t\|_2,
\end{equation}
where $w_t^i$ indicates the weights of local model of $i-th$ client. This metric measures the distance between a model $w_t^i$ and the nearest model of other clients. Note that $d(i)$ is equal to $0$ if $i-th$ client is clean. Considering the bound of $d(i)$, it is further normalized to $[0, 1]$ as $D(i)=\frac{d(i)}{\max_{j}d(j)}$. Then, local models are aggregated to update the global model by
\begin{equation}
    w^{t+1}=\sum_{i=1}^{|S_t|}\frac{n_ie^{-D(i)}}{\sum_{j=1}^{|S_t|}n_je^{-D(j)}}w_i^t.
\end{equation}
In this way, the aggregation weight of any clean client is constant due to $D(i) = 0$, while the weights of noisy clients are multiplied by a scaling factor in $[0, 1]$. Moreover, the weights would decrease as the model distance increases, adaptively adjusting the contributions of noisy clients in global model updating \cite{FedNoRo}.
% Similar intuition is in accordance with the early attempt in robust federated learning method krum \cite{krum} which explicitly computes the distances among different updated models. Similar idea is exploited in RSCFed, which aims to improve the aggregation robustness in the federated semi-supervised learning \cite{rscfed} .

\section{Experiments}
\label{sec:exp}
\subsection{Experimental Setup}
\label{setup}
% We list basic experimental settings as follows. 

\textbf{Datasets} 
In this study, we consider three types of label noise (detailedly discussed in Section \ref{sec:labelnoise}) and conduct experiments on five datasets and their statistics are shown in Table \ref{tab:dataset}. We normalize the images by the dataset's overall mean and standard deviation \cite{nguyen}. For data heterogeneity, we utilize Dirichlet distribution to distribute  total data to different clients, where each client is assigned with training data partitioned from the Dirichlet distribution with a concentration parameter $\gamma$, where a lower $\gamma$ indicates a higher Non-IID degree \cite{qinbinCrossSilo}. We set $\gamma$ to $1.0$ and $0.5$ to generate two Non-IID degree. For CIFAR-N datasets, we additionaly conduct experiments with IID partitioning, since it is less explored in previous works.
\begin{table}[htb]
  \centering
    \caption{Datasets and base models utilized in experiments. $\dagger$ denotes the pre-trained ResNet \cite{kaiming} on ImageNet \cite{deng2009imagenet} is used.}
  \label{tab:dataset}
  \begin{adjustbox}{width=\columnwidth,center}
\begin{tabular}{c|c|c|c}
\hline 
\textbf{Dataset}                      & CIFAR-10 \& CIFAR-10-N        & CIFAR-100 \& CIFAR-100-N                               & Clothing1M       \\ \hline
\textbf{Size of $\mathcal{D}$}        & 50,000          & 50,000                   & 1,000,000        \\
\textbf{Size of $\mathcal{D}_{test}$} & 10,000                       & 10,000              & 10,000           \\
\textbf{Classes $M$}      & 10                             & 100                 & 14               \\
\textbf{Base Model}                         & ResNet-18            & ResNet-34           & ResNet-50 $\dagger$        \\ \hline 
\end{tabular}
  \end{adjustbox}

\end{table}

\textbf{Baselines} 
%following the same method used in FedAvg.
For main experiments, we implement sixteen state-of-the-art methods (discussed in Section \ref{RW} and Table \ref{tab:comparison_prev}). Unless otherwise specified, most hyperparameters of these methods are configured favorably in line with the original literature. We refer to official open-source codes of these methods. For convenience, we clarify some important hyper-parameter settings as follows:
\begin{itemize}[leftmargin=0.1cm]
    \item General FL methods: FedAvg \cite{fedavg},  FedProx \cite{fedprox} and FedExP \cite{FedExP}. The $\mu$ of FedProx is fixed to 0.1 and $\epsilon$ of FedExP is fixed to $1e^{-3}$. 
    % Inspired by \cite{clipping,peijian}, we also evaluate FedAvg with gradient clipping since gradient clipping may provide robustness against overfitting local noisy dataset. Detailedly, the clipping norm threshold of local model's gradient is mildly set as $5.0$ as default, which means if a parameter's gradient for updating is higher than $5.0$ or lower than $-5.0$, it will be modified to $5$ or $-5$.
    \item Robust aggregation methods: Krum \cite{krum}, Median \cite{median} and TrimmededMean \cite{TrimmedMean}. Krum and TrimmedMean require to set the upper bound of the ratio of compromised or bad clients (denoted as $\kappa$), empirically  $\kappa$ is selected from $0.1-0.3$ \cite{Byzantium}. We select $\kappa$ to $0.3$ since  $\kappa$ is actually agnostic since different participants hold unknown extent of label noise in their datasets. 
    \item Noisy label learning (NLL) methods:  Joint Optim \cite{jointopt}, Symmetric CE \cite{symmetricCE}, SELFIE \cite{selfie}, Co-teaching \cite{co-teaching}, Co-teaching+ \cite{co-teaching+} and DivideMix \cite{Dividemix}. These methods are easy to combine with classic FedAvg's pipeline. The $\alpha$ and $\beta$ are set to 0.1 and $1.0$ for Symmetric CE. We report the average metrics of two peer networks of DivideMix, Co-teaching and Co-teaching+.
    \item Federated noisy label learning (FNLL) methods: Robust FL \cite{robustfl}, FedLSR \cite{FedLSR}, FedRN \cite{fedrn} and FedNoRo \cite{FedNoRo}. The warm-up epochs for RobustFL, FedLSR, FedRN are set to 20\% of total communication rounds. The $\gamma_e$ and $\gamma$ of FedLSR are set to $0.3$ and $0.4$. The reliable neighbor number is set as 1 for FedRN. The forget rate of Co-teaching, Co-teaching+ and RobustFL is mildly set to $0.2$.
\end{itemize}

\textbf{Implementation details}
All main experiments are implemented by Pytorch \cite{paszke2017automatic} on Nvidia$^\circledR$ RTX 3090 GPUs. We utilize class-wise averaged precision rate (Pre.) and recall rate (Rec.) as metrics for evaluation with the assistance of scikit-learn package \cite{sklearn}. All experiments are averaged over 3 seeds for a fair comparison. Mixed precision training \cite{mixedPrecision} is used to accelerate the model training. We generate $N=100$ clients in total, and 10 clients (i.e. $|S_t|=10$) participate in FL in each $t-th$ communication round, which is a vanilla experimental setting in FL \cite{fedavg,FedLSR}. We select the local iteration epoch $E$ of each client to 5, and local batchsize to 64. The SGD optimizer is selected with a fixed learning rate of $0.01$, momentum of $0.9$ and weight decay of $5e^{-4}$ referring to \cite{flexifed,decorr}. Total global round number is 120 for each method to achieve its convergence.

\textbf{Hyper-parameter selection} There are four hyper-parameters including the trade-off coefficients $\alpha$ and $\beta$ in Eq. \ref{eq:all_loss}, warm-up epochs $T_w$ of stage\#1 and the learning rate $\eta$ to update the possible label distribution $\tilde{y}$. We fix $\alpha$ and $\beta$ to $0.2$ and $0.5$ respectively. The warm-up epoch $T_w$ is selected as 20. The $\eta$ is fixed to $1000$ for CIFAR-10, CIFAR-10-N and Clothing1M datasets, and it is fixed to $5000$ for CIFAR-100 and CIFAR-100-N datasets referring to \cite{pencil}.

\subsection{Analysis on Synthetic Noisy Datasets}
\label{sec:cifar-m}
 We list the experimental results of synthetic noisy datasets with manually injected noisy labels in Table \ref{tab:results_c10} and Table \ref{tab:results_c100}. We generate two Non-IID settings along with four label noise scenarios as discussed in Section \ref{sec:labelnoise}. For noisy CIFAR-10 datasets, we observe our FedELC achieves the overall best performance while keeping a good balance between the precision rate and the recall rate, while other methods mainly exhibit evidently higher precision rate than the recall rate. Under the higher label noise scenario (i.e. symmetric label noise across $0.0-0.8$), FedLSR \cite{FedLSR} and Symmetric CE \cite{symmetricCE} have slight improvement over our FedELC in the aspect of the precision rate.  We also find robust aggregation methods seem to be not effective considering all settings. For example, Krum \cite{krum} is mainly designed against the byzantine issue in FL, which only selects one robust enough client-side updated model as the next-round global model and it fails to absorb enough information from other clients.
 
 For noisy CIFAR-100 datasets, our method still have overall competitive performance compared with these baseline methods when the total class $M$ is high. Additionally, we have some extra observations including \textbf{(1)} Co-teaching \cite{co-teaching}, which maintains two peer networks, is also a robust training method considering different label noise settings. \textbf{(2)} General FL methods have surprisingly robust performance for most settings. Notably, compared with other curated methods, these methods have simple training scheme and less hyper-parameters. This indicates an underlying possibility where we can base on the simple general FL methods and train robust model in FL with only simple techniques like \cite{unleashing}. It also indicates for clients with lower noise rates, we can use some vanilla training methods like Eq. \ref{eq:cls} (refer to Section \ref{sec:stage2} and Algorithm \ref{algo}) to conduct local update. \textbf{(3)} Some FNLL methods get evident performance degrades when the total class $M$ get higher. For Robust FL \cite{robustfl}, we reckon that it relies on the global class-wise class centroids to form significant supervision for local updates, while this technique faces difficulties since the data are both Non-IID and of label noise. For FedLSR \cite{FedLSR}, we reckon the degrade of FedLSR lies in the distillation based method is not suitable when the total class number $M$ is high, which has also been analyzed in previous works \cite{lsg}.

\begin{table*}[htbp]
\caption{Precision (Pre.) and recall rates (Rec.) on synthetic noisy dataset CIFAR-10 with manually-injected noisy labels. Sym./Asym. refer to symmetric/asymmetric label noise. The \textbf{bold} and \underline{underlined} denote the best and second-place result.}
\label{tab:results_c10}
\resizebox{\linewidth}{!}{
\begin{tabular}{c|cccccccc|cccccccc}
\hline
\multirow{3}{*}{\textbf{Method}} & \multicolumn{8}{c|}{Dirichlet ($\gamma$=1.0)} & \multicolumn{8}{c}{Dirichlet ($\gamma$=0.5)} \\ \cline{2-17} 
 & \multicolumn{2}{c|}{Sym. (0.0-0.4)} & \multicolumn{2}{c|}{Sym. (0.0-0.8)} & \multicolumn{2}{c|}{Asym. (0.0-0.4)} & \multicolumn{2}{c|}{Mixed (0.0-0.4)} & \multicolumn{2}{c|}{Sym. (0.0-0.4)} & \multicolumn{2}{c|}{Sym. (0.0-0.8)} & \multicolumn{2}{c|}{Asym. (0.0-0.4)} & \multicolumn{2}{c}{Mixed (0.0-0.4)} \\
 & Pre. & \multicolumn{1}{c|}{Rec.} & Pre. & \multicolumn{1}{c|}{Rec.} & Pre. & \multicolumn{1}{c|}{Rec.} & Pre. & Rec. & Pre. & \multicolumn{1}{c|}{Rec.} & Pre. & \multicolumn{1}{c|}{Rec.} & Pre. & \multicolumn{1}{c|}{Rec.} & Pre. & Rec. \\ \hline
FedAvg \cite{fedavg} & 75.85 & \multicolumn{1}{c|}{73.58} & 57.41 & \multicolumn{1}{c|}{54.74} & 77.60 & \multicolumn{1}{c|}{76.08} & 77.54 & \underline{75.72} & 72.11 & \multicolumn{1}{c|}{62.98} & 49.44 & \multicolumn{1}{c|}{45.34} & 74.29 & \multicolumn{1}{c|}{64.66} & 73.76 & 67.55 \\
FedProx \cite{fedprox}& 73.03 & \multicolumn{1}{c|}{71.40} & 55.69 & \multicolumn{1}{c|}{53.81} & 77.41 & \multicolumn{1}{c|}{76.18} & 77.44 & 75.62 & 69.50 & \multicolumn{1}{c|}{63.41} & 47.95 & \multicolumn{1}{c|}{43.89} & 74.49 & \multicolumn{1}{c|}{64.68} & 73.86 & 67.85 \\
FedExp \cite{FedExP}& 75.81 & \multicolumn{1}{c|}{73.36} & 57.35 & \multicolumn{1}{c|}{54.68} & 77.40 & \multicolumn{1}{c|}{75.99} & 77.39 & 75.41 & 72.24 & \multicolumn{1}{c|}{62.98} & 50.21 & \multicolumn{1}{c|}{45.56} & 74.29 & \multicolumn{1}{c|}{64.25} & 73.82 & 67.90 \\
TrimmedMean \cite{TrimmedMean}& 71.21 & \multicolumn{1}{c|}{64.29} & 47.93 & \multicolumn{1}{c|}{44.41} & 73.06 & \multicolumn{1}{c|}{66.06} & 70.77 & 64.71 & 68.94 & \multicolumn{1}{c|}{60.96} & 49.97 & \multicolumn{1}{c|}{42.90} & 69.44 & \multicolumn{1}{c|}{57.17} & 71.86 & 59.70 \\
Krum \cite{krum}& 51.90 & \multicolumn{1}{c|}{40.41} & 39.60 & \multicolumn{1}{c|}{33.59} & 48.29 & \multicolumn{1}{c|}{42.68} & 43.28 & 48.54 & 33.49 & \multicolumn{1}{c|}{36.28} & 25.19 & \multicolumn{1}{c|}{28.45} & 18.54 & \multicolumn{1}{c|}{24.77} & 25.12 & 28.38 \\
Median \cite{median}& 72.52 & \multicolumn{1}{c|}{70.51} & 58.45 & \multicolumn{1}{c|}{56.56} & 75.56 & \multicolumn{1}{c|}{72.67} & 73.60 & 71.53 & 68.45 & \multicolumn{1}{c|}{63.79} & 52.30 & \multicolumn{1}{c|}{48.23} & 72.06 & \multicolumn{1}{c|}{64.85} & 72.09 & 65.93 \\
Co-teaching \cite{co-teaching}& 75.79 & \multicolumn{1}{c|}{73.52} & 57.30 & \multicolumn{1}{c|}{55.03} & 76.88 & \multicolumn{1}{c|}{74.77} & 75.37 & 73.10 & 72.23 & \multicolumn{1}{c|}{66.62} & 48.23 & \multicolumn{1}{c|}{43.93} & \underline{74.95} & \multicolumn{1}{c|}{64.48} & 74.22 & 68.18 \\
Co-teaching+ \cite{co-teaching}& 69.70 & \multicolumn{1}{c|}{65.47} & 64.71 & \multicolumn{1}{c|}{59.07} & 67.60 & \multicolumn{1}{c|}{61.74} & 63.70 & 58.07 & 58.09 & \multicolumn{1}{c|}{46.97} & 57.01 & \multicolumn{1}{c|}{44.73} & 46.79 & \multicolumn{1}{c|}{41.31} & 55.72 & 44.69 \\
Joint Optim \cite{jointopt}& 64.74 & \multicolumn{1}{c|}{64.69} & 59.78 & \multicolumn{1}{c|}{59.55} & 65.02 & \multicolumn{1}{c|}{64.77} & 64.53 & 64.43 & 57.76 & \multicolumn{1}{c|}{57.37} & 52.44 & \multicolumn{1}{c|}{52.18} & 55.21 & \multicolumn{1}{c|}{54.34} & 57.84 & 57.12 \\
SELFIE \cite{selfie}& \underline{76.70} & \multicolumn{1}{c|}{\underline{74.55}} & 65.30 & \multicolumn{1}{c|}{59.51} & 77.84 & \multicolumn{1}{c|}{73.80} & \underline{77.55} & 74.74 & \underline{72.60} & \multicolumn{1}{c|}{64.46} & 60.08 & \multicolumn{1}{c|}{50.30} & 73.36 & \multicolumn{1}{c|}{60.10} & \underline{74.26} & 64.13 \\
Symmetric CE \cite{symmetricCE} & 76.32 & \multicolumn{1}{c|}{73.38} & \textbf{70.96} & \multicolumn{1}{c|}{\underline{66.25}} & 74.46 & \multicolumn{1}{c|}{70.94} & 76.13 & 72.83 & 72.05 & \multicolumn{1}{c|}{61.45} & \textbf{62.39}& \multicolumn{1}{c|}{54.64} & 70.93 & \multicolumn{1}{c|}{58.37} & 71.48 & 58.39 \\
DivideMix \cite{Dividemix}& 63.28 & \multicolumn{1}{c|}{61.94} & 60.19 & \multicolumn{1}{c|}{58.98} & 65.36 & \multicolumn{1}{c|}{65.30} & 66.78 & 65.40 & 57.36 & \multicolumn{1}{c|}{50.37} & 55.09 & \multicolumn{1}{c|}{50.21} & 63.03 & \multicolumn{1}{c|}{61.93} & 62.23 & 58.34 \\ \hline
Robust FL \cite{robustfl}& 63.17 & \multicolumn{1}{c|}{61.68} & 51.34 & \multicolumn{1}{c|}{44.67} & 63.26 & \multicolumn{1}{c|}{61.80} & 62.65 & 60.65 & 54.05 & \multicolumn{1}{c|}{45.06} & 55.45 & \multicolumn{1}{c|}{24.37} & 54.08 & \multicolumn{1}{c|}{42.86} & 53.94 & 43.41 \\
FedLSR \cite{FedLSR}& 75.16 & \multicolumn{1}{c|}{71.22} & \underline{70.42} & \multicolumn{1}{c|}{66.23} & 75.11 & \multicolumn{1}{c|}{71.29} & 73.34 & 68.52 & 64.20 & \multicolumn{1}{c|}{57.54} & \underline{62.35}& \multicolumn{1}{c|}{55.97} & 66.46 & \multicolumn{1}{c|}{55.82} & 60.04 & 53.12 \\
FedRN \cite{fedrn}& 72.65 & \multicolumn{1}{c|}{67.96} & 62.89 & \multicolumn{1}{c|}{55.65} & 69.02 & \multicolumn{1}{c|}{67.19} & 72.23 & 61.03 & 58.72 & \multicolumn{1}{c|}{46.46} & 54.47 & \multicolumn{1}{c|}{50.14} & 54.14 & \multicolumn{1}{c|}{40.29} & 55.76 & 47.81 \\
FedNoRo \cite{FedNoRo}& 73.67 & \multicolumn{1}{c|}{73.52} & 66.50 & \multicolumn{1}{c|}{66.18} & \underline{77.88} & \multicolumn{1}{c|}{\underline{76.77}} & 75.38 & 75.02 & 71.09 & \multicolumn{1}{c|}{\underline{71.01}} & 59.11 & \multicolumn{1}{c|}{\underline{57.41}} & 73.63 & \multicolumn{1}{c|}{\underline{72.77}} & 73.51 & \underline{72.35} \\ \hline
FedELC &  \textbf{76.81}& \multicolumn{1}{c|}{\textbf{76.72}} & 68.22 & \multicolumn{1}{c|}{\textbf{66.61}} & \textbf{77.97} & \multicolumn{1}{c|}{\textbf{76.98}} & \textbf{77.98} & \textbf{77.24} & \textbf{73.13} & \multicolumn{1}{c|}{\textbf{71.31}} & 60.31 & \multicolumn{1}{c|}{\textbf{59.67}} & \textbf{75.78} & \multicolumn{1}{c|}{\textbf{74.65}} & \textbf{74.55}  & \textbf{73.80} \\ \hline
\end{tabular}
}
\end{table*}

\begin{table}[htbp]
\caption{Experimental results (\%) on CIFAR-10-N dataset. }
\label{tab:cifar10-n}
\small
\begin{adjustbox}{width=\columnwidth,center}
\begin{tabular}{c|cc|cccc}
\hline
\multirow{3}{*}{\textbf{Method}} & \multicolumn{2}{c|}{IID} & \multicolumn{4}{c}{Non-IID} \\ \cline{2-7} 
 & \multicolumn{2}{c|}{-} & \multicolumn{2}{c|}{Dir. ($\gamma=1.0$)} & \multicolumn{2}{c}{Dir. ($\gamma=0.5$)} \\
 & Pre. & Rec. & Pre. & \multicolumn{1}{c|}{Rec.} & Pre. & Rec. \\ \hline
FedAvg \cite{fedavg}& \underline{87.42} & \underline{87.25} & 83.85 & \multicolumn{1}{c|}{82.79} & 81.37 & 73.40 \\
FedProx \cite{fedprox} & 86.79 & 86.77 & \underline{83.97} & \multicolumn{1}{c|}{83.50} & 81.31 & 77.09 \\
FedExP \cite{FedExP}& 87.36 & 87.31 & 83.88 & \multicolumn{1}{c|}{83.71} & 81.22 & 73.70 \\
TrimmedMean \cite{TrimmedMean}& 86.02 & 85.88 & 81.05 & \multicolumn{1}{c|}{78.02} & 75.75 & 65.72 \\
Krum \cite{krum}& 74.58 & 72.56 & 68.61 & \multicolumn{1}{c|}{61.29} & 44.16 & 36.84 \\
Median \cite{median} & 86.10 & 86.03 & 83.03 & \multicolumn{1}{c|}{81.47} & 78.47 & 71.82 \\
Co-teaching \cite{co-teaching}& 84.61 & 84.55 & 79.76 & \multicolumn{1}{c|}{77.05} & 73.12 & 61.59 \\
Co-teaching+ \cite{co-teaching+}& 78.68 & 78.37 & 71.08 & \multicolumn{1}{c|}{65.12} & 60.12 & 50.78 \\
Joint Optim \cite{jointopt}& 78.22 & 78.35 & 68.09 & \multicolumn{1}{c|}{68.03} & 59.97 & 59.72 \\
SELFIE \cite{selfie}& 83.79 & 83.72 & 78.73 & \multicolumn{1}{c|}{74.52} & 7366 & 58.34 \\
Symmetric CE \cite{symmetricCE}& 83.84 & 83.65 & 78.42 & \multicolumn{1}{c|}{75.47} & 72.67 & 62.06 \\
DivideMix \cite{Dividemix}& 76.77 & 75.22 & 66.29 & \multicolumn{1}{c|}{66.02} & 59.32 & 56.07 \\ \hline
Robust FL \cite{robustfl}& 75.77 & 75.58 & 65.33 & \multicolumn{1}{c|}{63.71} & 55.35 & 51.88 \\
FedLSR \cite{FedLSR}& 82.24 & 81.95 & 77.35 & \multicolumn{1}{c|}{74.29} & 60.78 & 53.82 \\
FedRN \cite{fedrn}& 75.61 & 75.10 & 76.45 & \multicolumn{1}{c|}{70.87} & 47.30 & 46.20 \\
FedNoRo \cite{FedNoRo}& 87.24 & 87.20 & 83.90 & \multicolumn{1}{c|}{\underline{83.81}} & \underline{82.87} & \underline{81.68} \\ \hline
FedELC & \textbf{87.67} & \textbf{87.47} & \textbf{84.29}  & \multicolumn{1}{c|}{\textbf{84.19}} & \textbf{83.36} & \textbf{82.25} \\ \hline
\end{tabular}
\end{adjustbox}
\end{table}

\begin{table*}[htbp]
\caption{Experimental results (\%) on synthetic noisy dataset CIFAR-100 with manually-injected noisy labels.}
\label{tab:results_c100}
\resizebox{\linewidth}{!}{
\begin{tabular}{c|cccccccc|cccccccc}
\hline 
\multirow{3}{*}{\textbf{Method}} & \multicolumn{8}{c|}{Dirichlet ($\gamma$=1.0)} & \multicolumn{8}{c}{Dirichlet ($\gamma$=0.5)} \\ \cline{2-17} 
 & \multicolumn{2}{c|}{Sym. (0.0-0.4)} & \multicolumn{2}{c|}{Sym. (0.0-0.8)} & \multicolumn{2}{c|}{Asym. (0.0-0.4)} & \multicolumn{2}{c|}{Mixed (0.0-0.4)} & \multicolumn{2}{c|}{Sym. (0.0-0.4)} & \multicolumn{2}{c|}{Sym. (0.0-0.8)} & \multicolumn{2}{c|}{Asym. (0.0-0.4)} & \multicolumn{2}{c}{Mixed (0.0-0.4)} \\
 & Pre. & \multicolumn{1}{c|}{Rec.} & Pre. & \multicolumn{1}{c|}{Rec.} & Pre. & \multicolumn{1}{c|}{Rec.} & Pre. & Rec. & Pre. & \multicolumn{1}{c|}{Rec.} & Pre. & \multicolumn{1}{c|}{Rec.} & Pre. & \multicolumn{1}{c|}{Rec.} & Pre. & Rec. \\ \hline
FedAvg \cite{fedavg}& 43.51 & \multicolumn{1}{c|}{41.01} & 31.31 & \multicolumn{1}{c|}{28.38} & 43.57& \multicolumn{1}{c|}{41.52} & 42.06 & 40.37 & 42.11& \multicolumn{1}{c|}{40.49} & 30.27 & \multicolumn{1}{c|}{27.23} & 44.42& \multicolumn{1}{c|}{43.36} & 41.84& 40.74 \\
FedProx \cite{fedprox}& 40.82 & \multicolumn{1}{c|}{38.27} & 29.16 & \multicolumn{1}{c|}{27.33} & 44.33 & \multicolumn{1}{c|}{41.76} & 42.98 & 39.92 & 40.48 & \multicolumn{1}{c|}{37.34} & 27.49 & \multicolumn{1}{c|}{25.17} & 44.30& \multicolumn{1}{c|}{42.56} & 42.66& 39.59 \\
FedExP \cite{FedExP}& 43.88 & \multicolumn{1}{c|}{40.36} & 31.26 & \multicolumn{1}{c|}{28.23} & 44.46& \multicolumn{1}{c|}{42.62} & 41.74 & 40.27 & 42.74& \multicolumn{1}{c|}{40.46} & 29.80 & \multicolumn{1}{c|}{26.93} & 44.91& \multicolumn{1}{c|}{\underline{43.57}} & 42.53& 40.79\\
TrimmedMean \cite{TrimmedMean}& 41.44 & \multicolumn{1}{c|}{35.44} & 28.33 & \multicolumn{1}{c|}{24.50} & 44.46 & \multicolumn{1}{c|}{41.13} & 41.48 & 35.99 & 41.56 & \multicolumn{1}{c|}{34.50} & 22.95 & \multicolumn{1}{c|}{19.01} & 44.16 & \multicolumn{1}{c|}{37.20} & 42.45& 36.11 \\
Krum \cite{krum}& 22.99 & \multicolumn{1}{c|}{17.34} & 17.36 & \multicolumn{1}{c|}{14.08} & 24.88 & \multicolumn{1}{c|}{19.36} & 24.32 & 18.61 & 14.40 & \multicolumn{1}{c|}{11.92} & 7.95 & \multicolumn{1}{c|}{7.55} & 13.12 & \multicolumn{1}{c|}{11.35} & 12.29 & 11.04 \\
Median \cite{median}& 43.90 & \multicolumn{1}{c|}{34.47} & 31.01& \multicolumn{1}{c|}{22.24} & 43.20 & \multicolumn{1}{c|}{37.63} & 43.44 & 37.29 & 44.59 & \multicolumn{1}{c|}{28.79} & 30.05& \multicolumn{1}{c|}{19.27} & 50.47 & \multicolumn{1}{c|}{33.07} & 40.31& 31.93 \\
Co-teaching \cite{co-teaching}& \underline{44.74}& \multicolumn{1}{c|}{42.53} & 32.72 & \multicolumn{1}{c|}{30.86} & 44.78& \multicolumn{1}{c|}{42.28} &\textbf{45.64}& \underline{42.68} & \textbf{47.83} & \multicolumn{1}{c|}{\underline{43.01}} & 30.04& \multicolumn{1}{c|}{29.74} & \textbf{45.52} & \multicolumn{1}{c|}{43.48} & \textbf{45.37}& \underline{42.73}\\
Co-teaching+ \cite{co-teaching+}& 31.03 & \multicolumn{1}{c|}{33.63} & 24.09 & \multicolumn{1}{c|}{27.71} & 31.35 & \multicolumn{1}{c|}{33.38} & 31.91 & 33.67 & 28.77 & \multicolumn{1}{c|}{31.47} & 22.80 & \multicolumn{1}{c|}{27.16} & 30.35 & \multicolumn{1}{c|}{31.87} & 30.26 & 31.73 \\
Joint Optim \cite{jointopt}& 27.66 & \multicolumn{1}{c|}{28.67} & 22.67 & \multicolumn{1}{c|}{23.59} & 27.84 & \multicolumn{1}{c|}{29.11} & 27.83 & 28.88 & 26.39 & \multicolumn{1}{c|}{27.45} & 21.95 & \multicolumn{1}{c|}{22.80} & 26.20 & \multicolumn{1}{c|}{27.37} & 26.30 & 27.50 \\
SELFIE \cite{symmetricCE}& 44.62 & \multicolumn{1}{c|}{42.92} & 32.66 & \multicolumn{1}{c|}{30.59} & \textbf{44.90} & \multicolumn{1}{c|}{42.87} & 42.77& 41.18& 45.06 & \multicolumn{1}{c|}{42.26} & 31.82 & \multicolumn{1}{c|}{29.31} & 46.36 & \multicolumn{1}{c|}{42.54} & 42.20& 41.95\\
Symmetric CE \cite{symmetricCE}& 42.35 & \multicolumn{1}{c|}{39.64} & 32.11& \multicolumn{1}{c|}{\underline{32.86}} & 42.44 & \multicolumn{1}{c|}{38.82} & 42.41 & 39.93 & 40.99 & \multicolumn{1}{c|}{37.28} & 31.64& \multicolumn{1}{c|}{30.65} & 40.90 & \multicolumn{1}{c|}{37.03} & 42.05 & 38.18 \\
DivideMix \cite{Dividemix}& 37.42 & \multicolumn{1}{c|}{37.68} & 30.74 & \multicolumn{1}{c|}{31.04} & 36.78 & \multicolumn{1}{c|}{37.13} & 38.59 & 39.01 & 37.23 & \multicolumn{1}{c|}{37.39} & 31.79 & \multicolumn{1}{c|}{31.87} & 36.98 & \multicolumn{1}{c|}{37.26} & 36.86 & 37.01 \\ \hline
Robust FL \cite{robustfl}& 17.41 & \multicolumn{1}{c|}{10.69} & 3.73 & \multicolumn{1}{c|}{5.02} & 14.67 & \multicolumn{1}{c|}{9.03} & 17.59 & 9.45 & 11.23 & \multicolumn{1}{c|}{7.82} & 5.03 & \multicolumn{1}{c|}{5.71} & 10.50 & \multicolumn{1}{c|}{7.44} & 13.54 & 8.19 \\
FedLSR \cite{FedLSR}& 3.08 & \multicolumn{1}{c|}{11.46} & 2.80 & \multicolumn{1}{c|}{10.71} & 3.14 & \multicolumn{1}{c|}{11.39} & 3.55 & 11.48 & 3.84 & \multicolumn{1}{c|}{12.07} & 3.18 & \multicolumn{1}{c|}{10.40} & 3.56 & \multicolumn{1}{c|}{11.08} & 3.66 & 11.49 \\
FedRN \cite{fedrn}& 32.48 & \multicolumn{1}{c|}{26.73} & 26.35 & \multicolumn{1}{c|}{21.70} & 35.22 & \multicolumn{1}{c|}{27.81} & 35.72 & 26.19 & 38.04 & \multicolumn{1}{c|}{25.98} & 29.78 & \multicolumn{1}{c|}{19.48} & 36.73 & \multicolumn{1}{c|}{25.67} & 34.39 & 23.47 \\
FedNoRo \cite{FedNoRo}& 44.70 & \multicolumn{1}{c|}{\underline{43.44}} & \underline{32.98} & \multicolumn{1}{c|}{32.01} & \underline{44.51} & \multicolumn{1}{c|}{\underline{43.56}} & 43.32& 42.51 & 45.09 & \multicolumn{1}{c|}{40.48} & \textbf{33.79} & \multicolumn{1}{c|}{\underline{32.58}} & 43.97 & \multicolumn{1}{c|}{42.84} & 43.54& 41.92\\ \hline
FedELC & \textbf{44.83} & \multicolumn{1}{c|}{\textbf{43.65}} & \textbf{33.07} & \multicolumn{1}{c|}{\textbf{32.95}} & 44.01& \multicolumn{1}{c|}{\textbf{43.83}} &  \underline{43.87}&  \textbf{42.73}& \underline{45.22} & \multicolumn{1}{c|}{\textbf{43.12}} & \underline{32.98}  & \multicolumn{1}{c|}{\textbf{32.67}} &  \underline{44.97}& \multicolumn{1}{c|}{\textbf{43.64}} &  \underline{43.59}&  \textbf{42.81}\\ \hline
\end{tabular}
}
\end{table*}

% \usepackage{multirow}

% \usepackage{multirow}
\begin{table}[htbp]
\caption{Experimental results (\%) on CIFAR-100-N dataset.}
\label{tab:cifar100-n}
\begin{adjustbox}{width=\columnwidth,center}
\begin{tabular}{c|cc|cccc}
\hline
\multirow{3}{*}{\textbf{Method}} & \multicolumn{2}{c|}{IID} & \multicolumn{4}{c}{Non-IID} \\ \cline{2-7} 
 & \multicolumn{2}{c|}{-} & \multicolumn{2}{c|}{Dir. ($\gamma=1.0$)} & \multicolumn{2}{c}{Dir. ($\gamma=0.5$)} \\
 & Pre. & Rec. & Pre. & \multicolumn{1}{c|}{Rec.} & Pre. & Rec. \\ \hline
FedAvg \cite{fedavg}& 56.89& \underline{56.37} & \textbf{58.19}& \multicolumn{1}{c|}{\textbf{55.98}} & \underline{57.95}& \underline{54.27} \\
FedProx \cite{fedprox}& 56.22 & 53.33& 54.73& \multicolumn{1}{c|}{53.38} & 55.92& 54.01\\
FedExP \cite{FedExP}& 56.80& 55.99& \underline{57.33} & \multicolumn{1}{c|}{54.54} & \textbf{58.41}& \textbf{55.01}\\
TrimmedMean \cite{TrimmedMean}& 54.02 & 52.50 & 56.10 & \multicolumn{1}{c|}{49.59} & 56.29 & 48.30 \\
Krum \cite{krum}& 39.39 & 35.60 & 35.93 & \multicolumn{1}{c|}{23.55} & 21.71 & 18.61 \\
Median \cite{median}& 54.77 & 50.63 & 55.95 & \multicolumn{1}{c|}{47.87} & 55.32 & 44.65 \\
Co-teaching \cite{co-teaching}& 56.97 & 56.04 & 57.19 & \multicolumn{1}{c|}{\underline{55.56}} & 56.61 & 54.04 \\
Co-teaching+ \cite{co-teaching+}& 37.75 & 39.12 & 35.52 & \multicolumn{1}{c|}{36.95} & 32.48 & 33.96 \\
Joint Optim \cite{jointopt}& 33.43 & 34.37 & 31.93 & \multicolumn{1}{c|}{32.95} & 30.34 & 31.33 \\
SELFIE \cite{selfie}& 5646 & 55.47 & 57.61 & \multicolumn{1}{c|}{55.46} & 57.11 & 53.83 \\
Symmetric CE \cite{symmetricCE}& 46.63 & 45.46 & 46.34 & \multicolumn{1}{c|}{43.16} & 42.26 & 38.64 \\
DivideMix \cite{Dividemix}& 41.70 & 42.46 & 43.76 & \multicolumn{1}{c|}{44.66} & 45.25 & 45.70 \\ \hline
Robust FL \cite{robustfl}& 33.08 & 18.33 & 29.59 & \multicolumn{1}{c|}{16.72} & 21.39 & 12.49 \\
FedLSR \cite{FedLSR}& 5.07 & 14.61 & 4.23 & \multicolumn{1}{c|}{13.88} & 5.51 & 13.51 \\
FedRN \cite{fedrn}& 35.49 & 33.96 & 38.02 & \multicolumn{1}{c|}{32.13} & 35.49 & 26.57 \\
FedNoRo \cite{FedNoRo}& \underline{57.04} & 56.30 & 54.24 & \multicolumn{1}{c|}{53.30} & 53.42 & 51.02 \\ \hline
FedELC & \textbf{57.13} & \textbf{56.54} & 54.33 & \multicolumn{1}{c|}{53.32} & 53.17 & 53.07 \\ \hline
\end{tabular}
\end{adjustbox}
\end{table}

%Co-teaching has a hyperparameter $R(T)$ which cannot be priorly inferred in the practical federated learning system (discussed in section \ref{results}).

\subsection{Analysis on Datasets with Real-world Human Annotation Errors}
\label{sec:humanoise}

Now we analyze the evaluation results from the datasets containing real-world human annotation label noise patterns. The experimental results are listed in Table \ref{tab:cifar10-n} and Table \ref{tab:cifar100-n}. For the CIFAR-10-N dataset, we find our method FedELC shows stably better performance than other baseline methods. Comparing Table \ref{tab:cifar10-n} and Table \ref{tab:results_c10} for datasets with manual label noise, we find that comparing with the label noise issue caused by human-annotation errors, manual injected label noise exists as an harder challenge for learning with noisy labels since nearly all methods have evidently higher performance than all settings of manual-injected label noise settings.

For CIFAR-100-N, we get similar observations indicating manual label noise is harder than the label noise caused by human annotation errors. Our method shows better performance on IID partitioned data while it gets slight performance degrades on Non-IID partitioned data. We also find the general FL methods including FedAvg, FedExP and FedProx almost show superior performance than other NLL and FNLL methods, which is an interesting observation. This also indicates that simpler method can have satisfying performance against the human annotation caused label noise.

\subsection{Sensitivity \& Ablation Study}
\label{sec:ablation}
Herein we elaborate on the robustness of FedELC by conducting ablation study on hyper-parameter selection and two utilized techniques in our framework. Experimental settings are in accordance with the setting of CIFAR-10 of manual label noise ($0.0-0.4$).

\begin{figure}[hbp]
    \centering
    \includegraphics[width=\linewidth]{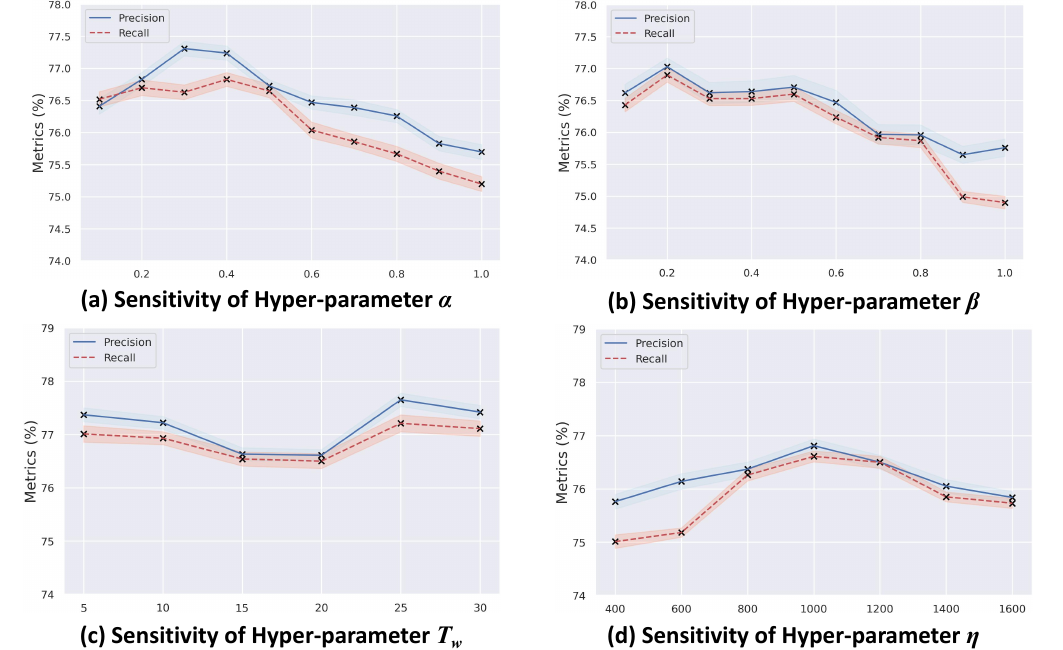}
    \caption{Experimental results for sensitivity study.}
    \label{fig:sensitivity}
\end{figure}

\textbf{Sensitivity on the hyper-parameter selection} We study the sensitivity of introduced hyper-parameters and the experimental results are shown in Figure \ref{fig:sensitivity}. From these results, we can observe that: \textbf{(1)} Our method has robust performance on the selection of  the trade-off coefficients $\alpha$ and $\beta$.  The recommended selection is that $\alpha$ and $\beta$ are smaller than 0.5.   \textbf{(2) } A larger $T_w$ can lead to higher final performance. \textbf{(3)} Our method has robust performance on the selection of $\eta$ while a mild learning rate $\eta$ for $\tilde{y}$ updating can lead to a slightly better performance. A small $\eta$ yield a slight performance degrade and perhaps it is because a small $\eta$ leads a slow update of the $\tilde{y}$ and also decreases the training efficiency. For guidance, when applying to other datasets, it is suggested to observe the gradient value of $\tilde{y}$ to select the suitable $\eta$ which can effectively update $\tilde{y}$. 

\textbf{Ablation on the logit adjustment \& model aggregation}
We conduct ablation experiments in Table \ref{tab:ablation_on_trick} by removing the logit adjustment technique \cite{LA} discussed in Eq. \ref{eq:cls} and DA aggregation technique \cite{FedNoRo}. The performance get degraded by removing each, which reflects the effectiveness of these incorporated techniques.

\begin{table}[htbp]
\caption{Results for Ablation study.}
\label{tab:ablation_on_trick}
\begin{adjustbox}{width=\columnwidth,center}
\begin{tabular}{cc|cccc}
\hline
\multicolumn{2}{c|}{Component} & \multicolumn{4}{c}{Metric} \\ \hline
\multicolumn{1}{c|}{Logit Adjustment} & DA Aggregation & \multicolumn{1}{c|}{Pre.} & \multicolumn{1}{c|}{Rec.} & \multicolumn{1}{c|}{@F1-Score} & @Acc. \\ \hline
\multicolumn{1}{c|}{$\checkmark$} & $\checkmark$ & \multicolumn{1}{c|}{76.81} & \multicolumn{1}{c|}{76.72} & \multicolumn{1}{c|}{76.78} & 77.03 \\ \hline
\multicolumn{1}{c|}{$\checkmark$} & $\times$ & \multicolumn{1}{c|}{76.44} & \multicolumn{1}{c|}{76.28} & \multicolumn{1}{c|}{76.36} &  76.43\\ \hline
\multicolumn{1}{c|}{$\times$} & $\checkmark$ & \multicolumn{1}{c|}{75.50} & \multicolumn{1}{c|}{75.33} & \multicolumn{1}{c|}{75.41} &  75.56\\ \hline
\multicolumn{1}{c|}{$\times$} & $\times$ & \multicolumn{1}{c|}{74.89} & \multicolumn{1}{c|}{74.77} & \multicolumn{1}{c|}{74.83} &  74.91\\ \hline
\end{tabular}
\end{adjustbox}
\end{table}

\subsection{Analysis on Noisy Client Detection}
\label{sec:detection}
\begin{figure}[tb]
    \centering
    \includegraphics[width=\linewidth]{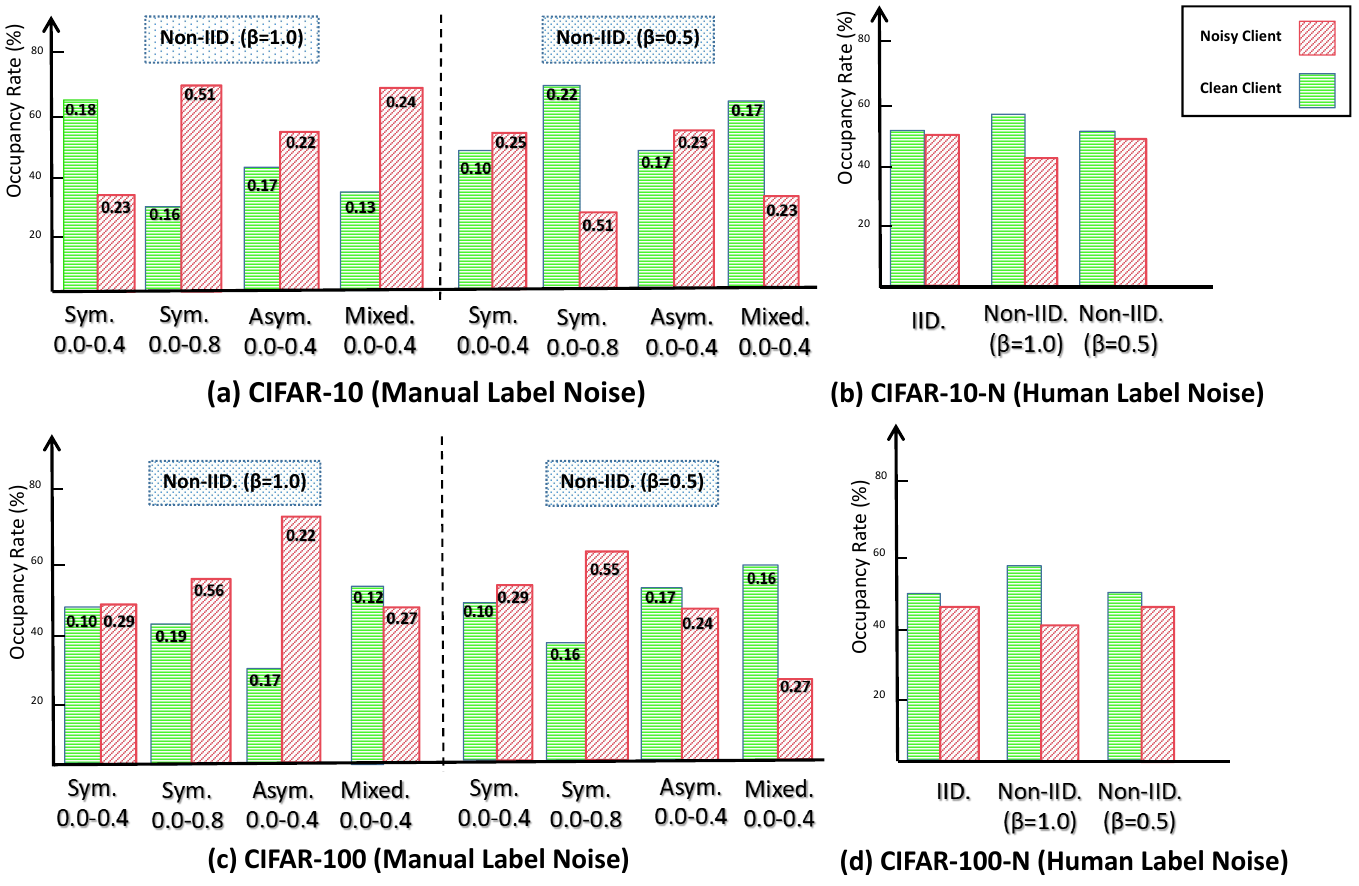}
    \caption{Visualization on the client division.}
    \label{fig:division}
\end{figure}

Here we visualize the detection and average noise rates within detected clean group and noisy group in Figure \ref{fig:division}. Experiments are repeated for five times. For datasets containing manual label noise, we can compute the average noise rate within each group, since we linearly generate the label noise rate according to the client index (e.g., for symmetric label noise scenario, client$\#$1 has all clean labels while client$\#100$ has the biggest noise rate). From these results, this division can relatively divide the total clients into the relatively clean group and the relatively noisy group, and the noisy group has higher averaged noise rate than the clean group. For datasets containing human-annotation errors, we find the clean group has a bit more clients than the noisy group while clients of both groups have similar noise rate (about 40\% as discussed in Section \ref{sec:labelnoise}).

\subsection{Analysis on End-to-end Label Correction}
\label{sec:correction}
We reckon considering noisy labels in FL, we can not only improve the trained model's robustness against label noise but also try to conduct label correction. In Figure \ref{fig:elc}, we visualize the label correction performance of FedELC and Joint Optimization framework \cite{jointopt} after $T_w$ global rounds for manual label noise scenarios, which are harder than human label noise as analyzed in Section \ref{sec:humanoise}. We use our estimated label $y_{estimate}$ in Algorithm \ref{algo} and compare it with the true ground-truth label. We backup the true labels before adding label noise to the dataset to conveniently compute this correction accuracy. The results demonstrate our method shows higher label estimation accuracy than another label correction method. Additionally, participants of FL can take fewer efforts to compare the original labels $\hat{y}$ with our estimated label $y_{estimate}$ and re-label some inconsistent samples ($\hat{y} \neq y_{estimate}$) to further improve the local data quality, which saves a lot of human efforts.

\begin{figure}[tb]
    \centering
    \includegraphics[width=\linewidth]{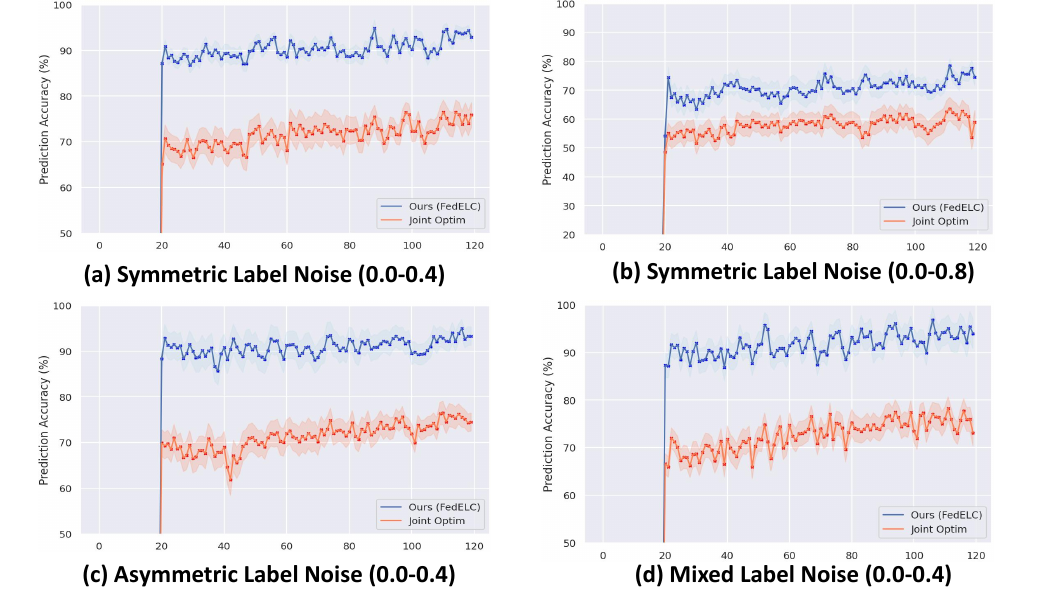}
    \caption{Label correction performance curves.}
    \label{fig:elc}
\end{figure}

\subsection{Analysis on large-scale Clothing1M dataset}
For preprocessing, all images are resized to $224\times224$ and normalized. 
The global communication lasts for 40 rounds. 
Results are shown in Table \ref{tab:testonclothing}. 
We additionally evaluate FedCorr \cite{fedcorr} and ELR \cite{ELR} following the open-source codes.
For \#15, learning rate is fixed to 0.1 to achieve higher performance. 
For Robust FL \cite{robustfl} and FedNoRo \cite{FedNoRo}, warm-up period lasts 10 global rounds. 
Experimental results verify combining the fine-grained noisy client identification and online end-to-end label correction, our method FedELC achieves more reliable performance against its counterparts without compromising the privacy. Robust FL shows higher performance in the complicated systematic label noise scenarios, which indicates the potential of utilizing the class-wise feature centroids as extra supervision to robustly regularize the model training (refer to Section \ref{sec:fnll}). We can introduce similar ideas into our method to further enhance the robustness, which we leave for future directions.

\begin{table}[tbp]
\caption{Results on systematic noisy dataset Clothing1M with random partitioning. $\dagger$ denotes RobustFL requires exchanging the sensitive information as discussed in \cite{FedLSR,robustfl,fedcorr}.}
\label{tab:testonclothing}
\small
% \resizebox{\linewidth}{!}{
\centering
\begin{tabular}{ccc}
\hline
\# & \textbf{Method}                 & Best Test Accuracy (\%)\\ \hline
1           & FedAvg \cite{fedavg}                & 69.26\\ 
2           & FedProx \cite{fedprox}                & 67.62\\ 
% 3           & FedSAM                         & 68.94            & 69.84                           \\ \hline
3           & FedExP \cite{FedExP}      & 69.53 \\ 
4           & TrimmedMean \cite{TrimmedMean}                        & 68.57\\ 
5           & Krum \cite{krum}      & 67.32 \\ 
6           & Median \cite{median}                   & 69.00 \\ 
7           & Co-teaching \cite{co-teaching} & 69.77 \\ 
8           & Co-teaching+ \cite{co-teaching+}      & 68.02 \\ 
9           & Joint Optim \cite{jointopt}      & 70.31\\ 
10           & SELFIE \cite{selfie}                  & 69.97\\ 
11           & Symmetric CE \cite{symmetricCE}                 & 69.61 \\  
12           & DivideMix \cite{Dividemix}                        & 70.08\\ 
13          & ELR \cite{ELR}      & 68.66 \\ 
14           & RobustFL $\dagger$ \cite{robustfl}      & \textbf{71.77}\\ 
15           & FedLSR \cite{FedLSR}                        & 69.76\\ 
16           & FedRN \cite{fedrn}      & 68.60  \\ 
17           & FedCorr \cite{fedcorr}      & 69.92 \\
18           & FedNoRo \cite{FedNoRo}      & 70.52 \\\hline
19            & FedELC       & \underline{71.64} \\  \hline
\end{tabular}
% }
\end{table}

\section{Discussion \& Conclusion}
\label{conclude}

% \usepackage{multirow}

%In this paper, we have considered that each local dataset may consist of noisy labels, which cause serious performance degradation in the practical federated learning system. 
% We first discuss the limitations of this work and the possible future works to expand this work.

In this work, we focus on improving robustness against noisy data in federated learning (FL), which is an underlying prominent challenge when deploying a practical FL system. 
We propose a two-stage framework FedELC to detect the clients with possible higher noise levels, and also refine the unreliable labels of these noisy clients via an end-to-end correction manner.
One side effect of the proposed FedELC framework lies in it requires extra computation (about $\times0.3$ extra wall-clock  time for local training) to simultaneously conduct local training and end-to-end label correction for detected noisy clients. 
Extensive experiments demonstrate FedELC's robustness against sixteen baseline methods. We showcase the effectiveness of FedELC to correct local labels. Meanwhile, noisy clients can compare our given corrected labels with original labels to refine local datasets with less human efforts which further improves the quality of local data.
For future works, we aim to inspect model pruning \cite{biad}, gradient clipping \cite{peijian}, contribution estimation \cite{vldbCo,icdeFa}, generating reliable data \cite{fd} and other regularization techniques \cite{unleashing,sasvd,locNforget} for federated noisy label learning and also improve robustness against noisy data in more applications \cite{naacl,recomm}.

\section{Acknowledgment}
This work was supported by the National Key Research and Development Program of China (2021YFB2900102) and the National Natural Science Foundation of China (No. 62072436).

\bibliographystyle{ACM-Reference-Format}
\bibliography{sample-base}


\begin{thebibliography}{77}

%%% the Web address.

\ifx \showCODEN    \undefined \def \showCODEN     #1{\unskip}     \fi
\ifx \showDOI      \undefined \def \showDOI       #1{#1}\fi
\ifx \showISBNx    \undefined \def \showISBNx     #1{\unskip}     \fi
\ifx \showISBNxiii \undefined \def \showISBNxiii  #1{\unskip}     \fi
\ifx \showISSN     \undefined \def \showISSN      #1{\unskip}     \fi
\ifx \showLCCN     \undefined \def \showLCCN      #1{\unskip}     \fi
\ifx \shownote     \undefined \def \shownote      #1{#1}          \fi
\ifx \showarticletitle \undefined \def \showarticletitle #1{#1}   \fi
\ifx \showURL      \undefined \def \showURL       {\relax}        \fi
% invisible to TeX
\providecommand\bibfield[2]{#2}
\providecommand\bibinfo[2]{#2}
\providecommand\natexlab[1]{#1}
\providecommand\showeprint[2][]{arXiv:#2}

\bibitem[Allen-Zhu and Li(2022)]%
        {allenZhu}
\bibfield{author}{\bibinfo{person}{Zeyuan Allen-Zhu} {and} \bibinfo{person}{Yuanzhi Li}.} \bibinfo{year}{2022}\natexlab{}.
\newblock \showarticletitle{Towards Understanding Ensemble, Knowledge Distillation and Self-Distillation in Deep Learning}. In \bibinfo{booktitle}{\emph{The Eleventh International Conference on Learning Representations}}.
\newblock


\bibitem[Arpit et~al\mbox{.}(2017)]%
        {memorization}
\bibfield{author}{\bibinfo{person}{Devansh Arpit}, \bibinfo{person}{Stanislaw Jastrzebski}, \bibinfo{person}{Nicolas Ballas}, \bibinfo{person}{David Krueger}, \bibinfo{person}{Emmanuel Bengio}, \bibinfo{person}{Maxinder~S. Kanwal}, \bibinfo{person}{Tegan Maharaj}, \bibinfo{person}{Asja Fischer}, \bibinfo{person}{Aaron~C. Courville}, \bibinfo{person}{Yoshua Bengio}, {and} \bibinfo{person}{Simon Lacoste{-}Julien}.} \bibinfo{year}{2017}\natexlab{}.
\newblock \showarticletitle{A Closer Look at Memorization in Deep Networks}. In \bibinfo{booktitle}{\emph{Proceedings of the 34th International Conference on Machine Learning, {ICML} 2017, Sydney, NSW, Australia, 6-11 August 2017}} \emph{(\bibinfo{series}{Proceedings of Machine Learning Research}, Vol.~\bibinfo{volume}{70})}, \bibfield{editor}{\bibinfo{person}{Doina Precup} {and} \bibinfo{person}{Yee~Whye Teh}} (Eds.). \bibinfo{publisher}{{PMLR}}, \bibinfo{pages}{233--242}.
\newblock
\urldef\tempurl%
\url{http://proceedings.mlr.press/v70/arpit17a.html}
\showURL{%
\tempurl}


\bibitem[Bai et~al\mbox{.}(2023)]%
        {lsg}
\bibfield{author}{\bibinfo{person}{Daokuan Bai}, \bibinfo{person}{Shanshan Wang}, \bibinfo{person}{Wenyue Wang}, \bibinfo{person}{Hua Wang}, \bibinfo{person}{Chuan Zhao}, \bibinfo{person}{Peng Yuan}, {and} \bibinfo{person}{Zhenxiang Chen}.} \bibinfo{year}{2023}\natexlab{}.
\newblock \showarticletitle{Overcoming Noisy Labels in Federated Learning Through Local Self-Guiding}. In \bibinfo{booktitle}{\emph{23rd {IEEE/ACM} International Symposium on Cluster, Cloud and Internet Computing, CCGrid 2023, Bangalore, India, May 1-4, 2023}}, \bibfield{editor}{\bibinfo{person}{Yogesh Simmhan}, \bibinfo{person}{Ilkay Altintas}, \bibinfo{person}{Ana~Lucia Varbanescu}, \bibinfo{person}{Pavan Balaji}, \bibinfo{person}{Abhinandan~S. Prasad}, {and} \bibinfo{person}{Lorenzo Carnevale}} (Eds.). \bibinfo{publisher}{{IEEE}}, \bibinfo{pages}{367--376}.
\newblock
\urldef\tempurl%
\url{https://doi.org/10.1109/CCGrid57682.2023.00042}
\showDOI{\tempurl}


\bibitem[Berthelot et~al\mbox{.}(2019)]%
        {mixmatch}
\bibfield{author}{\bibinfo{person}{David Berthelot}, \bibinfo{person}{Nicholas Carlini}, \bibinfo{person}{Ian~J. Goodfellow}, \bibinfo{person}{Nicolas Papernot}, \bibinfo{person}{Avital Oliver}, {and} \bibinfo{person}{Colin Raffel}.} \bibinfo{year}{2019}\natexlab{}.
\newblock \showarticletitle{MixMatch: {A} Holistic Approach to Semi-Supervised Learning}. In \bibinfo{booktitle}{\emph{Advances in Neural Information Processing Systems 32: Annual Conference on Neural Information Processing Systems 2019, NeurIPS 2019, December 8-14, 2019, Vancouver, BC, Canada}}, \bibfield{editor}{\bibinfo{person}{Hanna~M. Wallach}, \bibinfo{person}{Hugo Larochelle}, \bibinfo{person}{Alina Beygelzimer}, \bibinfo{person}{Florence d'Alch{\'{e}}{-}Buc}, \bibinfo{person}{Emily~B. Fox}, {and} \bibinfo{person}{Roman Garnett}} (Eds.). \bibinfo{pages}{5050--5060}.
\newblock
\urldef\tempurl%
\url{https://proceedings.neurips.cc/paper/2019/hash/1cd138d0499a68f4bb72bee04bbec2d7-Abstract.html}
\showURL{%
\tempurl}


\bibitem[Blanchard et~al\mbox{.}(2017)]%
        {krum}
\bibfield{author}{\bibinfo{person}{Peva Blanchard}, \bibinfo{person}{El~Mahdi~El Mhamdi}, \bibinfo{person}{Rachid Guerraoui}, {and} \bibinfo{person}{Julien Stainer}.} \bibinfo{year}{2017}\natexlab{}.
\newblock \showarticletitle{Machine Learning with Adversaries: Byzantine Tolerant Gradient Descent}. In \bibinfo{booktitle}{\emph{Advances in Neural Information Processing Systems 30: Annual Conference on Neural Information Processing Systems 2017, December 4-9, 2017, Long Beach, CA, {USA}}}, \bibfield{editor}{\bibinfo{person}{Isabelle Guyon}, \bibinfo{person}{Ulrike von Luxburg}, \bibinfo{person}{Samy Bengio}, \bibinfo{person}{Hanna~M. Wallach}, \bibinfo{person}{Rob Fergus}, \bibinfo{person}{S.~V.~N. Vishwanathan}, {and} \bibinfo{person}{Roman Garnett}} (Eds.). \bibinfo{pages}{119--129}.
\newblock
\urldef\tempurl%
\url{https://proceedings.neurips.cc/paper/2017/hash/f4b9ec30ad9f68f89b29639786cb62ef-Abstract.html}
\showURL{%
\tempurl}


\bibitem[Chen et~al\mbox{.}(2022)]%
        {fedmc}
\bibfield{author}{\bibinfo{person}{Cen Chen}, \bibinfo{person}{Tiandi Ye}, \bibinfo{person}{Li Wang}, {and} \bibinfo{person}{Ming Gao}.} \bibinfo{year}{2022}\natexlab{}.
\newblock \showarticletitle{Learning to generalize in heterogeneous federated networks}. In \bibinfo{booktitle}{\emph{Proceedings of the 31st ACM International Conference on Information \& Knowledge Management}}. \bibinfo{pages}{159--168}.
\newblock


\bibitem[Chen et~al\mbox{.}(2024)]%
        {vldbCo}
\bibfield{author}{\bibinfo{person}{Yiwei Chen}, \bibinfo{person}{Kaiyu Li}, \bibinfo{person}{Guoliang Li}, {and} \bibinfo{person}{Yong Wang}.} \bibinfo{year}{2024}\natexlab{}.
\newblock \showarticletitle{Contributions Estimation in Federated Learning: A Comprehensive Experimental Evaluation}.
\newblock \bibinfo{journal}{\emph{Proceedings of the VLDB Endowment}} \bibinfo{volume}{17}, \bibinfo{number}{8} (\bibinfo{year}{2024}), \bibinfo{pages}{2077--2090}.
\newblock


\bibitem[Chen et~al\mbox{.}(2020)]%
        {chen2020focus}
\bibfield{author}{\bibinfo{person}{Yiqiang Chen}, \bibinfo{person}{Xiaodong Yang}, \bibinfo{person}{Xin Qin}, \bibinfo{person}{Han Yu}, \bibinfo{person}{Biao Chen}, {and} \bibinfo{person}{Zhiqi Shen}.} \bibinfo{year}{2020}\natexlab{}.
\newblock \showarticletitle{Focus: Dealing with label quality disparity in federated learning}.
\newblock \bibinfo{journal}{\emph{arXiv preprint arXiv:2001.11359}} (\bibinfo{year}{2020}).
\newblock


\bibitem[Deng et~al\mbox{.}(2009)]%
        {deng2009imagenet}
\bibfield{author}{\bibinfo{person}{Jia Deng}, \bibinfo{person}{Wei Dong}, \bibinfo{person}{Richard Socher}, \bibinfo{person}{Li-Jia Li}, \bibinfo{person}{Kai Li}, {and} \bibinfo{person}{Li Fei-Fei}.} \bibinfo{year}{2009}\natexlab{}.
\newblock \showarticletitle{Imagenet: A large-scale hierarchical image database}. In \bibinfo{booktitle}{\emph{2009 IEEE conference on computer vision and pattern recognition}}. Ieee, \bibinfo{pages}{248--255}.
\newblock


\bibitem[Fang and Ye(2022)]%
        {rhfl}
\bibfield{author}{\bibinfo{person}{Xiuwen Fang} {and} \bibinfo{person}{Mang Ye}.} \bibinfo{year}{2022}\natexlab{}.
\newblock \showarticletitle{Robust Federated Learning with Noisy and Heterogeneous Clients}. In \bibinfo{booktitle}{\emph{{IEEE/CVF} Conference on Computer Vision and Pattern Recognition, {CVPR} 2022, New Orleans, LA, USA, June 18-24, 2022}}. \bibinfo{publisher}{{IEEE}}, \bibinfo{pages}{10062--10071}.
\newblock
\urldef\tempurl%
\url{https://doi.org/10.1109/CVPR52688.2022.00983}
\showDOI{\tempurl}


\bibitem[Grandvalet and Bengio(2004)]%
        {entropyy}
\bibfield{author}{\bibinfo{person}{Yves Grandvalet} {and} \bibinfo{person}{Yoshua Bengio}.} \bibinfo{year}{2004}\natexlab{}.
\newblock \showarticletitle{Semi-supervised Learning by Entropy Minimization}. In \bibinfo{booktitle}{\emph{Advances in Neural Information Processing Systems 17 [Neural Information Processing Systems, {NIPS} 2004, December 13-18, 2004, Vancouver, British Columbia, Canada]}}. \bibinfo{pages}{529--536}.
\newblock
\urldef\tempurl%
\url{https://proceedings.neurips.cc/paper/2004/hash/96f2b50b5d3613adf9c27049b2a888c7-Abstract.html}
\showURL{%
\tempurl}


\bibitem[Han et~al\mbox{.}(2020)]%
        {hbsur}
\bibfield{author}{\bibinfo{person}{Bo Han}, \bibinfo{person}{Quanming Yao}, \bibinfo{person}{Tongliang Liu}, \bibinfo{person}{Gang Niu}, \bibinfo{person}{Ivor~W Tsang}, \bibinfo{person}{James~T Kwok}, {and} \bibinfo{person}{Masashi Sugiyama}.} \bibinfo{year}{2020}\natexlab{}.
\newblock \showarticletitle{A survey of label-noise representation learning: Past, present and future}.
\newblock \bibinfo{journal}{\emph{arXiv preprint arXiv:2011.04406}} (\bibinfo{year}{2020}).
\newblock


\bibitem[Han et~al\mbox{.}(2018)]%
        {co-teaching}
\bibfield{author}{\bibinfo{person}{Bo Han}, \bibinfo{person}{Quanming Yao}, \bibinfo{person}{Xingrui Yu}, \bibinfo{person}{Gang Niu}, \bibinfo{person}{Miao Xu}, \bibinfo{person}{Weihua Hu}, \bibinfo{person}{Ivor~W. Tsang}, {and} \bibinfo{person}{Masashi Sugiyama}.} \bibinfo{year}{2018}\natexlab{}.
\newblock \showarticletitle{Co-teaching: Robust training of deep neural networks with extremely noisy labels}. In \bibinfo{booktitle}{\emph{Advances in Neural Information Processing Systems 31: Annual Conference on Neural Information Processing Systems 2018, NeurIPS 2018, December 3-8, 2018, Montr{\'{e}}al, Canada}}, \bibfield{editor}{\bibinfo{person}{Samy Bengio}, \bibinfo{person}{Hanna~M. Wallach}, \bibinfo{person}{Hugo Larochelle}, \bibinfo{person}{Kristen Grauman}, \bibinfo{person}{Nicol{\`{o}} Cesa{-}Bianchi}, {and} \bibinfo{person}{Roman Garnett}} (Eds.). \bibinfo{pages}{8536--8546}.
\newblock
\urldef\tempurl%
\url{https://proceedings.neurips.cc/paper/2018/hash/a19744e268754fb0148b017647355b7b-Abstract.html}
\showURL{%
\tempurl}


\bibitem[He et~al\mbox{.}(2016)]%
        {kaiming}
\bibfield{author}{\bibinfo{person}{Kaiming He}, \bibinfo{person}{Xiangyu Zhang}, \bibinfo{person}{Shaoqing Ren}, {and} \bibinfo{person}{Jian Sun}.} \bibinfo{year}{2016}\natexlab{}.
\newblock \showarticletitle{Deep Residual Learning for Image Recognition}. In \bibinfo{booktitle}{\emph{2016 {IEEE} Conference on Computer Vision and Pattern Recognition, {CVPR} 2016, Las Vegas, NV, USA, June 27-30, 2016}}. \bibinfo{publisher}{{IEEE} Computer Society}, \bibinfo{pages}{770--778}.
\newblock
\urldef\tempurl%
\url{https://doi.org/10.1109/CVPR.2016.90}
\showDOI{\tempurl}


\bibitem[Hester et~al\mbox{.}(2018)]%
        {qlearning}
\bibfield{author}{\bibinfo{person}{Todd Hester}, \bibinfo{person}{Matej Vecerik}, \bibinfo{person}{Olivier Pietquin}, \bibinfo{person}{Marc Lanctot}, \bibinfo{person}{Tom Schaul}, \bibinfo{person}{Bilal Piot}, \bibinfo{person}{Dan Horgan}, \bibinfo{person}{John Quan}, \bibinfo{person}{Andrew Sendonaris}, \bibinfo{person}{Ian Osband}, {et~al\mbox{.}}} \bibinfo{year}{2018}\natexlab{}.
\newblock \showarticletitle{Deep q-learning from demonstrations}. In \bibinfo{booktitle}{\emph{Proceedings of the AAAI conference on artificial intelligence}}, Vol.~\bibinfo{volume}{32}.
\newblock


\bibitem[Hospedales et~al\mbox{.}(2021)]%
        {metalearning}
\bibfield{author}{\bibinfo{person}{Timothy Hospedales}, \bibinfo{person}{Antreas Antoniou}, \bibinfo{person}{Paul Micaelli}, {and} \bibinfo{person}{Amos Storkey}.} \bibinfo{year}{2021}\natexlab{}.
\newblock \showarticletitle{Meta-learning in neural networks: A survey}.
\newblock \bibinfo{journal}{\emph{IEEE transactions on pattern analysis and machine intelligence}} \bibinfo{volume}{44}, \bibinfo{number}{9} (\bibinfo{year}{2021}), \bibinfo{pages}{5149--5169}.
\newblock


\bibitem[Jhunjhunwala et~al\mbox{.}(2023)]%
        {FedExP}
\bibfield{author}{\bibinfo{person}{Divyansh Jhunjhunwala}, \bibinfo{person}{Shiqiang Wang}, {and} \bibinfo{person}{Gauri Joshi}.} \bibinfo{year}{2023}\natexlab{}.
\newblock \showarticletitle{FedExP: Speeding Up Federated Averaging via Extrapolation}. In \bibinfo{booktitle}{\emph{The Eleventh International Conference on Learning Representations, {ICLR} 2023, Kigali, Rwanda, May 1-5, 2023}}. \bibinfo{publisher}{OpenReview.net}.
\newblock
\urldef\tempurl%
\url{https://openreview.net/pdf?id=IPrzNbddXV}
\showURL{%
\tempurl}


\bibitem[Jiang et~al\mbox{.}(2022)]%
        {FedLSR}
\bibfield{author}{\bibinfo{person}{Xuefeng Jiang}, \bibinfo{person}{Sheng Sun}, \bibinfo{person}{Yuwei Wang}, {and} \bibinfo{person}{Min Liu}.} \bibinfo{year}{2022}\natexlab{}.
\newblock \showarticletitle{Towards Federated Learning against Noisy Labels via Local Self-Regularization}. In \bibinfo{booktitle}{\emph{Proceedings of the 31st {ACM} International Conference on Information {\&} Knowledge Management, Atlanta, GA, USA, October 17-21, 2022}}, \bibfield{editor}{\bibinfo{person}{Mohammad~Al Hasan} {and} \bibinfo{person}{Li~Xiong}} (Eds.). \bibinfo{publisher}{{ACM}}, \bibinfo{pages}{862--873}.
\newblock
\urldef\tempurl%
\url{https://doi.org/10.1145/3511808.3557475}
\showDOI{\tempurl}


\bibitem[Kang et~al\mbox{.}(2023)]%
        {unleashing}
\bibfield{author}{\bibinfo{person}{Hui Kang}, \bibinfo{person}{Sheng Liu}, \bibinfo{person}{Huaxi Huang}, \bibinfo{person}{Jun Yu}, \bibinfo{person}{Bo Han}, \bibinfo{person}{Dadong Wang}, {and} \bibinfo{person}{Tongliang Liu}.} \bibinfo{year}{2023}\natexlab{}.
\newblock \showarticletitle{Unleashing the Potential of Regularization Strategies in Learning with Noisy Labels}.
\newblock \bibinfo{journal}{\emph{CoRR}}  \bibinfo{volume}{abs/2307.05025} (\bibinfo{year}{2023}).
\newblock
\urldef\tempurl%
\url{https://doi.org/10.48550/arXiv.2307.05025}
\showDOI{\tempurl}
\showeprint[arXiv]{2307.05025}


\bibitem[Kim et~al\mbox{.}(2022)]%
        {fedrn}
\bibfield{author}{\bibinfo{person}{Sangmook Kim}, \bibinfo{person}{Wonyoung Shin}, \bibinfo{person}{Soohyuk Jang}, \bibinfo{person}{Hwanjun Song}, {and} \bibinfo{person}{Se{-}Young Yun}.} \bibinfo{year}{2022}\natexlab{}.
\newblock \showarticletitle{FedRN: Exploiting k-Reliable Neighbors Towards Robust Federated Learning}. In \bibinfo{booktitle}{\emph{Proceedings of the 31st {ACM} International Conference on Information {\&} Knowledge Management, Atlanta, GA, USA, October 17-21, 2022}}, \bibfield{editor}{\bibinfo{person}{Mohammad~Al Hasan} {and} \bibinfo{person}{Li~Xiong}} (Eds.). \bibinfo{publisher}{{ACM}}, \bibinfo{pages}{972--981}.
\newblock
\urldef\tempurl%
\url{https://doi.org/10.1145/3511808.3557322}
\showDOI{\tempurl}


\bibitem[Kim et~al\mbox{.}(2024)]%
        {flr}
\bibfield{author}{\bibinfo{person}{Taehyeon Kim}, \bibinfo{person}{Donggyu Kim}, {and} \bibinfo{person}{Se-Young Yun}.} \bibinfo{year}{2024}\natexlab{}.
\newblock \showarticletitle{Revisiting Early-Learning Regularization When Federated Learning Meets Noisy Labels}.
\newblock \bibinfo{journal}{\emph{arXiv preprint arXiv:2402.05353}} (\bibinfo{year}{2024}).
\newblock


\bibitem[Kuznetsova et~al\mbox{.}(2018)]%
        {machine-gene}
\bibfield{author}{\bibinfo{person}{Alina Kuznetsova}, \bibinfo{person}{Hassan Rom}, \bibinfo{person}{Neil Alldrin}, \bibinfo{person}{Jasper R.~R. Uijlings}, \bibinfo{person}{Ivan Krasin}, \bibinfo{person}{Jordi Pont{-}Tuset}, \bibinfo{person}{Shahab Kamali}, \bibinfo{person}{Stefan Popov}, \bibinfo{person}{Matteo Malloci}, \bibinfo{person}{Tom Duerig}, {and} \bibinfo{person}{Vittorio Ferrari}.} \bibinfo{year}{2018}\natexlab{}.
\newblock \showarticletitle{The Open Images Dataset {V4:} Unified image classification, object detection, and visual relationship detection at scale}.
\newblock \bibinfo{journal}{\emph{CoRR}}  \bibinfo{volume}{abs/1811.00982} (\bibinfo{year}{2018}).
\newblock
\showeprint[arXiv]{1811.00982}
\urldef\tempurl%
\url{http://arxiv.org/abs/1811.00982}
\showURL{%
\tempurl}


\bibitem[Lee et~al\mbox{.}(2022)]%
        {Byzantium}
\bibfield{author}{\bibinfo{person}{Youngjoon Lee}, \bibinfo{person}{Sangwoo Park}, {and} \bibinfo{person}{Joonhyuk Kang}.} \bibinfo{year}{2022}\natexlab{}.
\newblock \showarticletitle{Security-Preserving Federated Learning via Byzantine-Sensitive Triplet Distance}.
\newblock \bibinfo{journal}{\emph{CoRR}}  \bibinfo{volume}{abs/2210.16519} (\bibinfo{year}{2022}).
\newblock
\urldef\tempurl%
\url{https://doi.org/10.48550/arXiv.2210.16519}
\showDOI{\tempurl}
\showeprint[arXiv]{2210.16519}


\bibitem[Li et~al\mbox{.}(2024)]%
        {locNforget}
\bibfield{author}{\bibinfo{person}{Jia Li}, \bibinfo{person}{Lijie Hu}, \bibinfo{person}{Zhixian He}, \bibinfo{person}{Jingfeng Zhang}, \bibinfo{person}{Tianhang Zheng}, {and} \bibinfo{person}{Di Wang}.} \bibinfo{year}{2024}\natexlab{}.
\newblock \showarticletitle{Text Guided Image Editing with Automatic Concept Locating and Forgetting}.
\newblock \bibinfo{journal}{\emph{arXiv preprint arXiv:2405.19708}} (\bibinfo{year}{2024}).
\newblock


\bibitem[Li et~al\mbox{.}(2023)]%
        {fd}
\bibfield{author}{\bibinfo{person}{Jia Li}, \bibinfo{person}{Lijie Hu}, \bibinfo{person}{Jingfeng Zhang}, \bibinfo{person}{Tianhang Zheng}, \bibinfo{person}{Hua Zhang}, {and} \bibinfo{person}{Di Wang}.} \bibinfo{year}{2023}\natexlab{}.
\newblock \showarticletitle{Fair text-to-image diffusion via fair mapping}.
\newblock \bibinfo{journal}{\emph{arXiv preprint arXiv:2311.17695}} (\bibinfo{year}{2023}).
\newblock


\bibitem[Li et~al\mbox{.}(2022b)]%
        {peijian}
\bibfield{author}{\bibinfo{person}{Junyi Li}, \bibinfo{person}{Jian Pei}, {and} \bibinfo{person}{Heng Huang}.} \bibinfo{year}{2022}\natexlab{b}.
\newblock \showarticletitle{Communication-Efficient Robust Federated Learning with Noisy Labels}. In \bibinfo{booktitle}{\emph{{KDD} '22: The 28th {ACM} {SIGKDD} Conference on Knowledge Discovery and Data Mining, Washington, DC, USA, August 14 - 18, 2022}}, \bibfield{editor}{\bibinfo{person}{Aidong Zhang} {and} \bibinfo{person}{Huzefa Rangwala}} (Eds.). \bibinfo{publisher}{{ACM}}, \bibinfo{pages}{914--924}.
\newblock
\urldef\tempurl%
\url{https://doi.org/10.1145/3534678.3539328}
\showDOI{\tempurl}


\bibitem[Li et~al\mbox{.}(2020c)]%
        {Dividemix}
\bibfield{author}{\bibinfo{person}{Junnan Li}, \bibinfo{person}{Richard Socher}, {and} \bibinfo{person}{Steven C.~H. Hoi}.} \bibinfo{year}{2020}\natexlab{c}.
\newblock \showarticletitle{DivideMix: Learning with Noisy Labels as Semi-supervised Learning}. In \bibinfo{booktitle}{\emph{8th International Conference on Learning Representations, {ICLR} 2020, Addis Ababa, Ethiopia, April 26-30, 2020}}. \bibinfo{publisher}{OpenReview.net}.
\newblock
\urldef\tempurl%
\url{https://openreview.net/forum?id=HJgExaVtwr}
\showURL{%
\tempurl}


\bibitem[Li and Zhang(2024)]%
        {sasvd}
\bibfield{author}{\bibinfo{person}{Jia Li} {and} \bibinfo{person}{Hua Zhang}.} \bibinfo{year}{2024}\natexlab{}.
\newblock \showarticletitle{SA-SVD: Mitigating Bias in Face Recognition by Fair Representation Learning}. In \bibinfo{booktitle}{\emph{2024 27th International Conference on Computer Supported Cooperative Work in Design (CSCWD)}}. IEEE, \bibinfo{pages}{471--476}.
\newblock


\bibitem[Li et~al\mbox{.}(2022a)]%
        {qinbinCrossSilo}
\bibfield{author}{\bibinfo{person}{Qinbin Li}, \bibinfo{person}{Yiqun Diao}, \bibinfo{person}{Quan Chen}, {and} \bibinfo{person}{Bingsheng He}.} \bibinfo{year}{2022}\natexlab{a}.
\newblock \showarticletitle{Federated Learning on Non-IID Data Silos: An Experimental Study}. In \bibinfo{booktitle}{\emph{38th {IEEE} International Conference on Data Engineering, {ICDE} 2022, Kuala Lumpur, Malaysia, May 9-12, 2022}}. \bibinfo{publisher}{{IEEE}}, \bibinfo{pages}{965--978}.
\newblock
\urldef\tempurl%
\url{https://doi.org/10.1109/ICDE53745.2022.00077}
\showDOI{\tempurl}


\bibitem[Li et~al\mbox{.}(2021)]%
        {median}
\bibfield{author}{\bibinfo{person}{Tian Li}, \bibinfo{person}{Shengyuan Hu}, \bibinfo{person}{Ahmad Beirami}, {and} \bibinfo{person}{Virginia Smith}.} \bibinfo{year}{2021}\natexlab{}.
\newblock \showarticletitle{Ditto: Fair and Robust Federated Learning Through Personalization}. In \bibinfo{booktitle}{\emph{Proceedings of the 38th International Conference on Machine Learning, {ICML} 2021, 18-24 July 2021, Virtual Event}} \emph{(\bibinfo{series}{Proceedings of Machine Learning Research}, Vol.~\bibinfo{volume}{139})}, \bibfield{editor}{\bibinfo{person}{Marina Meila} {and} \bibinfo{person}{Tong Zhang}} (Eds.). \bibinfo{publisher}{{PMLR}}, \bibinfo{pages}{6357--6368}.
\newblock
\urldef\tempurl%
\url{http://proceedings.mlr.press/v139/li21h.html}
\showURL{%
\tempurl}


\bibitem[Li et~al\mbox{.}(2020a)]%
        {fedSummary}
\bibfield{author}{\bibinfo{person}{Tian Li}, \bibinfo{person}{Anit~Kumar Sahu}, \bibinfo{person}{Ameet Talwalkar}, {and} \bibinfo{person}{Virginia Smith}.} \bibinfo{year}{2020}\natexlab{a}.
\newblock \showarticletitle{Federated Learning: Challenges, Methods, and Future Directions}.
\newblock \bibinfo{journal}{\emph{{IEEE} Signal Process. Mag.}} \bibinfo{volume}{37}, \bibinfo{number}{3} (\bibinfo{year}{2020}), \bibinfo{pages}{50--60}.
\newblock
\urldef\tempurl%
\url{https://doi.org/10.1109/MSP.2020.2975749}
\showDOI{\tempurl}


\bibitem[Li et~al\mbox{.}(2020b)]%
        {fedprox}
\bibfield{author}{\bibinfo{person}{Tian Li}, \bibinfo{person}{Anit~Kumar Sahu}, \bibinfo{person}{Manzil Zaheer}, \bibinfo{person}{Maziar Sanjabi}, \bibinfo{person}{Ameet Talwalkar}, {and} \bibinfo{person}{Virginia Smith}.} \bibinfo{year}{2020}\natexlab{b}.
\newblock \showarticletitle{Federated Optimization in Heterogeneous Networks}. In \bibinfo{booktitle}{\emph{Proceedings of Machine Learning and Systems 2020, MLSys 2020, Austin, TX, USA, March 2-4, 2020}}, \bibfield{editor}{\bibinfo{person}{Inderjit~S. Dhillon}, \bibinfo{person}{Dimitris~S. Papailiopoulos}, {and} \bibinfo{person}{Vivienne Sze}} (Eds.). \bibinfo{publisher}{mlsys.org}.
\newblock
\urldef\tempurl%
\url{https://proceedings.mlsys.org/book/316.pdf}
\showURL{%
\tempurl}


\bibitem[Liang et~al\mbox{.}(2022)]%
        {rscfed}
\bibfield{author}{\bibinfo{person}{Xiaoxiao Liang}, \bibinfo{person}{Yiqun Lin}, \bibinfo{person}{Huazhu Fu}, \bibinfo{person}{Lei Zhu}, {and} \bibinfo{person}{Xiaomeng Li}.} \bibinfo{year}{2022}\natexlab{}.
\newblock \showarticletitle{Rscfed: Random sampling consensus federated semi-supervised learning}. In \bibinfo{booktitle}{\emph{Proceedings of the IEEE/CVF Conference on computer vision and pattern recognition}}. \bibinfo{pages}{10154--10163}.
\newblock


\bibitem[Liu et~al\mbox{.}(2023)]%
        {junxu}
\bibfield{author}{\bibinfo{person}{Junxu Liu}, \bibinfo{person}{Jian Lou}, \bibinfo{person}{Li Xiong}, {and} \bibinfo{person}{Xiaofeng Meng}.} \bibinfo{year}{2023}\natexlab{}.
\newblock \showarticletitle{Personalized Differentially Private Federated Learning without Exposing Privacy Budgets}. In \bibinfo{booktitle}{\emph{Proceedings of the 32nd ACM International Conference on Information and Knowledge Management}}. \bibinfo{pages}{4140--4144}.
\newblock


\bibitem[Liu et~al\mbox{.}(2020)]%
        {ELR}
\bibfield{author}{\bibinfo{person}{Sheng Liu}, \bibinfo{person}{Jonathan Niles{-}Weed}, \bibinfo{person}{Narges Razavian}, {and} \bibinfo{person}{Carlos Fernandez{-}Granda}.} \bibinfo{year}{2020}\natexlab{}.
\newblock \showarticletitle{Early-Learning Regularization Prevents Memorization of Noisy Labels}. In \bibinfo{booktitle}{\emph{Advances in Neural Information Processing Systems 33: Annual Conference on Neural Information Processing Systems 2020, NeurIPS 2020, December 6-12, 2020, virtual}}, \bibfield{editor}{\bibinfo{person}{Hugo Larochelle}, \bibinfo{person}{Marc'Aurelio Ranzato}, \bibinfo{person}{Raia Hadsell}, \bibinfo{person}{Maria{-}Florina Balcan}, {and} \bibinfo{person}{Hsuan{-}Tien Lin}} (Eds.).
\newblock
\urldef\tempurl%
\url{https://proceedings.neurips.cc/paper/2020/hash/ea89621bee7c88b2c5be6681c8ef4906-Abstract.html}
\showURL{%
\tempurl}


\bibitem[Luo et~al\mbox{.}(2021)]%
        {skel}
\bibfield{author}{\bibinfo{person}{Junyu Luo}, \bibinfo{person}{Jianlei Yang}, \bibinfo{person}{Xucheng Ye}, \bibinfo{person}{Xin Guo}, {and} \bibinfo{person}{Weisheng Zhao}.} \bibinfo{year}{2021}\natexlab{}.
\newblock \showarticletitle{Fedskel: efficient federated learning on heterogeneous systems with skeleton gradients update}. In \bibinfo{booktitle}{\emph{Proceedings of the 30th ACM international conference on information \& knowledge management}}. \bibinfo{pages}{3283--3287}.
\newblock


\bibitem[McLachlan and Rathnayake(2014)]%
        {gmm}
\bibfield{author}{\bibinfo{person}{Geoffrey~J. McLachlan} {and} \bibinfo{person}{Suren~I. Rathnayake}.} \bibinfo{year}{2014}\natexlab{}.
\newblock \showarticletitle{On the number of components in a Gaussian mixture model}.
\newblock \bibinfo{journal}{\emph{WIREs Data Mining Knowl. Discov.}} \bibinfo{volume}{4}, \bibinfo{number}{5} (\bibinfo{year}{2014}), \bibinfo{pages}{341--355}.
\newblock
\urldef\tempurl%
\url{https://doi.org/10.1002/WIDM.1135}
\showDOI{\tempurl}


\bibitem[McMahan et~al\mbox{.}(2017)]%
        {fedavg}
\bibfield{author}{\bibinfo{person}{Brendan McMahan}, \bibinfo{person}{Eider Moore}, \bibinfo{person}{Daniel Ramage}, \bibinfo{person}{Seth Hampson}, {and} \bibinfo{person}{Blaise~Ag{\"{u}}era y Arcas}.} \bibinfo{year}{2017}\natexlab{}.
\newblock \showarticletitle{Communication-Efficient Learning of Deep Networks from Decentralized Data}. In \bibinfo{booktitle}{\emph{Proceedings of the 20th International Conference on Artificial Intelligence and Statistics, {AISTATS} 2017, 20-22 April 2017, Fort Lauderdale, FL, {USA}}} \emph{(\bibinfo{series}{Proceedings of Machine Learning Research}, Vol.~\bibinfo{volume}{54})}, \bibfield{editor}{\bibinfo{person}{Aarti Singh} {and} \bibinfo{person}{Xiaojin~(Jerry) Zhu}} (Eds.). \bibinfo{publisher}{{PMLR}}, \bibinfo{pages}{1273--1282}.
\newblock
\urldef\tempurl%
\url{http://proceedings.mlr.press/v54/mcmahan17a.html}
\showURL{%
\tempurl}


\bibitem[Menon et~al\mbox{.}(2021)]%
        {LA}
\bibfield{author}{\bibinfo{person}{Aditya~Krishna Menon}, \bibinfo{person}{Sadeep Jayasumana}, \bibinfo{person}{Ankit~Singh Rawat}, \bibinfo{person}{Himanshu Jain}, \bibinfo{person}{Andreas Veit}, {and} \bibinfo{person}{Sanjiv Kumar}.} \bibinfo{year}{2021}\natexlab{}.
\newblock \showarticletitle{Long-tail learning via logit adjustment}. In \bibinfo{booktitle}{\emph{9th International Conference on Learning Representations, {ICLR} 2021, Virtual Event, Austria, May 3-7, 2021}}. \bibinfo{publisher}{OpenReview.net}.
\newblock
\urldef\tempurl%
\url{https://openreview.net/forum?id=37nvvqkCo5}
\showURL{%
\tempurl}


\bibitem[Micikevicius et~al\mbox{.}(2018)]%
        {mixedPrecision}
\bibfield{author}{\bibinfo{person}{Paulius Micikevicius}, \bibinfo{person}{Sharan Narang}, \bibinfo{person}{Jonah Alben}, \bibinfo{person}{Gregory Diamos}, \bibinfo{person}{Erich Elsen}, \bibinfo{person}{David Garcia}, \bibinfo{person}{Boris Ginsburg}, \bibinfo{person}{Michael Houston}, \bibinfo{person}{Oleksii Kuchaiev}, \bibinfo{person}{Ganesh Venkatesh}, {et~al\mbox{.}}} \bibinfo{year}{2018}\natexlab{}.
\newblock \showarticletitle{Mixed Precision Training}. In \bibinfo{booktitle}{\emph{International Conference on Learning Representations}}.
\newblock


\bibitem[Natarajan et~al\mbox{.}(2013)]%
        {NLLnara}
\bibfield{author}{\bibinfo{person}{Nagarajan Natarajan}, \bibinfo{person}{Inderjit~S. Dhillon}, \bibinfo{person}{Pradeep Ravikumar}, {and} \bibinfo{person}{Ambuj Tewari}.} \bibinfo{year}{2013}\natexlab{}.
\newblock \showarticletitle{Learning with Noisy Labels}. In \bibinfo{booktitle}{\emph{Advances in Neural Information Processing Systems 26: 27th Annual Conference on Neural Information Processing Systems 2013. Proceedings of a meeting held December 5-8, 2013, Lake Tahoe, Nevada, United States}}, \bibfield{editor}{\bibinfo{person}{Christopher J.~C. Burges}, \bibinfo{person}{L{\'{e}}on Bottou}, \bibinfo{person}{Zoubin Ghahramani}, {and} \bibinfo{person}{Kilian~Q. Weinberger}} (Eds.). \bibinfo{pages}{1196--1204}.
\newblock
\urldef\tempurl%
\url{https://proceedings.neurips.cc/paper/2013/hash/3871bd64012152bfb53fdf04b401193f-Abstract.html}
\showURL{%
\tempurl}


\bibitem[Nguyen et~al\mbox{.}(2022)]%
        {nguyen}
\bibfield{author}{\bibinfo{person}{John Nguyen}, \bibinfo{person}{Jianyu Wang}, \bibinfo{person}{Kshitiz Malik}, \bibinfo{person}{Maziar Sanjabi}, {and} \bibinfo{person}{Michael Rabbat}.} \bibinfo{year}{2022}\natexlab{}.
\newblock \showarticletitle{Where to Begin? On the Impact of Pre-Training and Initialization in Federated Learning}. In \bibinfo{booktitle}{\emph{The Eleventh International Conference on Learning Representations}}.
\newblock


\bibitem[Nishi et~al\mbox{.}(2021)]%
        {augdesc}
\bibfield{author}{\bibinfo{person}{Kento Nishi}, \bibinfo{person}{Yi Ding}, \bibinfo{person}{Alex Rich}, {and} \bibinfo{person}{Tobias H{\"{o}}llerer}.} \bibinfo{year}{2021}\natexlab{}.
\newblock \showarticletitle{Augmentation Strategies for Learning With Noisy Labels}. In \bibinfo{booktitle}{\emph{{IEEE} Conference on Computer Vision and Pattern Recognition, {CVPR} 2021, virtual, June 19-25, 2021}}. \bibinfo{publisher}{Computer Vision Foundation / {IEEE}}, \bibinfo{pages}{8022--8031}.
\newblock
\urldef\tempurl%
\url{https://doi.org/10.1109/CVPR46437.2021.00793}
\showDOI{\tempurl}


\bibitem[Paszke et~al\mbox{.}(2017)]%
        {paszke2017automatic}
\bibfield{author}{\bibinfo{person}{Adam Paszke}, \bibinfo{person}{Sam Gross}, \bibinfo{person}{Soumith Chintala}, \bibinfo{person}{Gregory Chanan}, \bibinfo{person}{Edward Yang}, \bibinfo{person}{Zachary DeVito}, \bibinfo{person}{Zeming Lin}, \bibinfo{person}{Alban Desmaison}, \bibinfo{person}{Luca Antiga}, {and} \bibinfo{person}{Adam Lerer}.} \bibinfo{year}{2017}\natexlab{}.
\newblock \showarticletitle{Automatic differentiation in pytorch}.
\newblock  (\bibinfo{year}{2017}).
\newblock


\bibitem[Patrini et~al\mbox{.}(2017)]%
        {making}
\bibfield{author}{\bibinfo{person}{Giorgio Patrini}, \bibinfo{person}{Alessandro Rozza}, \bibinfo{person}{Aditya~Krishna Menon}, \bibinfo{person}{Richard Nock}, {and} \bibinfo{person}{Lizhen Qu}.} \bibinfo{year}{2017}\natexlab{}.
\newblock \showarticletitle{Making Deep Neural Networks Robust to Label Noise: {A} Loss Correction Approach}. In \bibinfo{booktitle}{\emph{2017 {IEEE} Conference on Computer Vision and Pattern Recognition, {CVPR} 2017, Honolulu, HI, USA, July 21-26, 2017}}. \bibinfo{publisher}{{IEEE} Computer Society}, \bibinfo{pages}{2233--2241}.
\newblock
\urldef\tempurl%
\url{https://doi.org/10.1109/CVPR.2017.240}
\showDOI{\tempurl}


\bibitem[Pedregosa et~al\mbox{.}(2011)]%
        {sklearn}
\bibfield{author}{\bibinfo{person}{Fabian Pedregosa}, \bibinfo{person}{Ga{\"e}l Varoquaux}, \bibinfo{person}{Alexandre Gramfort}, \bibinfo{person}{Vincent Michel}, \bibinfo{person}{Bertrand Thirion}, \bibinfo{person}{Olivier Grisel}, \bibinfo{person}{Mathieu Blondel}, \bibinfo{person}{Peter Prettenhofer}, \bibinfo{person}{Ron Weiss}, \bibinfo{person}{Vincent Dubourg}, {et~al\mbox{.}}} \bibinfo{year}{2011}\natexlab{}.
\newblock \showarticletitle{Scikit-learn: Machine learning in Python}.
\newblock \bibinfo{journal}{\emph{the Journal of machine Learning research}}  \bibinfo{volume}{12} (\bibinfo{year}{2011}), \bibinfo{pages}{2825--2830}.
\newblock


\bibitem[Radford et~al\mbox{.}(2021)]%
        {clip}
\bibfield{author}{\bibinfo{person}{Alec Radford}, \bibinfo{person}{Jong~Wook Kim}, \bibinfo{person}{Chris Hallacy}, \bibinfo{person}{Aditya Ramesh}, \bibinfo{person}{Gabriel Goh}, \bibinfo{person}{Sandhini Agarwal}, \bibinfo{person}{Girish Sastry}, \bibinfo{person}{Amanda Askell}, \bibinfo{person}{Pamela Mishkin}, \bibinfo{person}{Jack Clark}, {et~al\mbox{.}}} \bibinfo{year}{2021}\natexlab{}.
\newblock \showarticletitle{Learning transferable visual models from natural language supervision}. In \bibinfo{booktitle}{\emph{International conference on machine learning}}. PMLR, \bibinfo{pages}{8748--8763}.
\newblock


\bibitem[Sharma et~al\mbox{.}(2022)]%
        {naacl}
\bibfield{author}{\bibinfo{person}{Rahul Sharma}, \bibinfo{person}{Anil Ramakrishna}, \bibinfo{person}{Ansel MacLaughlin}, \bibinfo{person}{Anna Rumshisky}, \bibinfo{person}{Jimit Majmudar}, \bibinfo{person}{Clement Chung}, \bibinfo{person}{Salman Avestimehr}, {and} \bibinfo{person}{Rahul Gupta}.} \bibinfo{year}{2022}\natexlab{}.
\newblock \showarticletitle{Federated Learning with Noisy User Feedback}. In \bibinfo{booktitle}{\emph{Proceedings of the 2022 Conference of the North American Chapter of the Association for Computational Linguistics: Human Language Technologies}}. \bibinfo{pages}{2726--2739}.
\newblock


\bibitem[Shi et~al\mbox{.}(2022)]%
        {decorr}
\bibfield{author}{\bibinfo{person}{Yujun Shi}, \bibinfo{person}{Jian Liang}, \bibinfo{person}{Wenqing Zhang}, \bibinfo{person}{Vincent Tan}, {and} \bibinfo{person}{Song Bai}.} \bibinfo{year}{2022}\natexlab{}.
\newblock \showarticletitle{Towards Understanding and Mitigating Dimensional Collapse in Heterogeneous Federated Learning}. In \bibinfo{booktitle}{\emph{The Eleventh International Conference on Learning Representations}}.
\newblock


\bibitem[Song et~al\mbox{.}(2019)]%
        {selfie}
\bibfield{author}{\bibinfo{person}{Hwanjun Song}, \bibinfo{person}{Minseok Kim}, {and} \bibinfo{person}{Jae{-}Gil Lee}.} \bibinfo{year}{2019}\natexlab{}.
\newblock \showarticletitle{{SELFIE:} Refurbishing Unclean Samples for Robust Deep Learning}. In \bibinfo{booktitle}{\emph{Proceedings of the 36th International Conference on Machine Learning, {ICML} 2019, 9-15 June 2019, Long Beach, California, {USA}}} \emph{(\bibinfo{series}{Proceedings of Machine Learning Research}, Vol.~\bibinfo{volume}{97})}, \bibfield{editor}{\bibinfo{person}{Kamalika Chaudhuri} {and} \bibinfo{person}{Ruslan Salakhutdinov}} (Eds.). \bibinfo{publisher}{{PMLR}}, \bibinfo{pages}{5907--5915}.
\newblock
\urldef\tempurl%
\url{http://proceedings.mlr.press/v97/song19b.html}
\showURL{%
\tempurl}


\bibitem[Song et~al\mbox{.}(2022)]%
        {NLL}
\bibfield{author}{\bibinfo{person}{Hwanjun Song}, \bibinfo{person}{Minseok Kim}, \bibinfo{person}{Dongmin Park}, \bibinfo{person}{Yooju Shin}, {and} \bibinfo{person}{Jae-Gil Lee}.} \bibinfo{year}{2022}\natexlab{}.
\newblock \showarticletitle{Learning from noisy labels with deep neural networks: A survey}.
\newblock \bibinfo{journal}{\emph{IEEE Transactions on Neural Networks and Learning Systems}} (\bibinfo{year}{2022}).
\newblock


\bibitem[Tahmasebian et~al\mbox{.}(2022)]%
        {robustfed}
\bibfield{author}{\bibinfo{person}{Farnaz Tahmasebian}, \bibinfo{person}{Jian Lou}, {and} \bibinfo{person}{Li Xiong}.} \bibinfo{year}{2022}\natexlab{}.
\newblock \showarticletitle{RobustFed: {A} Truth Inference Approach for Robust Federated Learning}. In \bibinfo{booktitle}{\emph{Proceedings of the 31st {ACM} International Conference on Information {\&} Knowledge Management, Atlanta, GA, USA, October 17-21, 2022}}, \bibfield{editor}{\bibinfo{person}{Mohammad~Al Hasan} {and} \bibinfo{person}{Li~Xiong}} (Eds.). \bibinfo{publisher}{{ACM}}, \bibinfo{pages}{1868--1877}.
\newblock
\urldef\tempurl%
\url{https://doi.org/10.1145/3511808.3557439}
\showDOI{\tempurl}


\bibitem[Tan et~al\mbox{.}(2020)]%
        {tanben}
\bibfield{author}{\bibinfo{person}{Ben Tan}, \bibinfo{person}{Bo Liu}, \bibinfo{person}{Vincent Zheng}, {and} \bibinfo{person}{Qiang Yang}.} \bibinfo{year}{2020}\natexlab{}.
\newblock \showarticletitle{A federated recommender system for online services}. In \bibinfo{booktitle}{\emph{Proceedings of the 14th ACM conference on recommender systems}}. \bibinfo{pages}{579--581}.
\newblock


\bibitem[Tanaka et~al\mbox{.}(2018)]%
        {jointopt}
\bibfield{author}{\bibinfo{person}{Daiki Tanaka}, \bibinfo{person}{Daiki Ikami}, \bibinfo{person}{Toshihiko Yamasaki}, {and} \bibinfo{person}{Kiyoharu Aizawa}.} \bibinfo{year}{2018}\natexlab{}.
\newblock \showarticletitle{Joint Optimization Framework for Learning With Noisy Labels}. In \bibinfo{booktitle}{\emph{2018 {IEEE} Conference on Computer Vision and Pattern Recognition, {CVPR} 2018, Salt Lake City, UT, USA, June 18-22, 2018}}. \bibinfo{publisher}{Computer Vision Foundation / {IEEE} Computer Society}, \bibinfo{pages}{5552--5560}.
\newblock
\urldef\tempurl%
\url{https://doi.org/10.1109/CVPR.2018.00582}
\showDOI{\tempurl}


\bibitem[Tao et~al\mbox{.}(2023)]%
        {tao2023local}
\bibfield{author}{\bibinfo{person}{Yingfan Tao}, \bibinfo{person}{Jingna Sun}, \bibinfo{person}{Hao Yang}, \bibinfo{person}{Li Chen}, \bibinfo{person}{Xu Wang}, \bibinfo{person}{Wenming Yang}, \bibinfo{person}{Daniel Du}, {and} \bibinfo{person}{Min Zheng}.} \bibinfo{year}{2023}\natexlab{}.
\newblock \showarticletitle{Local and Global Logit Adjustments for Long-Tailed Learning}. In \bibinfo{booktitle}{\emph{Proceedings of the IEEE/CVF International Conference on Computer Vision}}. \bibinfo{pages}{11783--11792}.
\newblock


\bibitem[Tsouvalas et~al\mbox{.}(2023)]%
        {chaos}
\bibfield{author}{\bibinfo{person}{Vasileios Tsouvalas}, \bibinfo{person}{Aaqib Saeed}, \bibinfo{person}{Tanir Ozcelebi}, {and} \bibinfo{person}{Nirvana Meratnia}.} \bibinfo{year}{2023}\natexlab{}.
\newblock \showarticletitle{Labeling Chaos to Learning Harmony: Federated Learning with Noisy Labels}.
\newblock \bibinfo{journal}{\emph{ACM Transactions on Intelligent Systems and Technology}} (\bibinfo{year}{2023}).
\newblock


\bibitem[Wang et~al\mbox{.}(2020)]%
        {wanghao}
\bibfield{author}{\bibinfo{person}{Hao Wang}, \bibinfo{person}{Zakhary Kaplan}, \bibinfo{person}{Di Niu}, {and} \bibinfo{person}{Baochun Li}.} \bibinfo{year}{2020}\natexlab{}.
\newblock \showarticletitle{Optimizing federated learning on non-iid data with reinforcement learning}. In \bibinfo{booktitle}{\emph{IEEE INFOCOM 2020-IEEE conference on computer communications}}. IEEE, \bibinfo{pages}{1698--1707}.
\newblock


\bibitem[Wang et~al\mbox{.}(2023a)]%
        {flexifed}
\bibfield{author}{\bibinfo{person}{Kaibin Wang}, \bibinfo{person}{Qiang He}, \bibinfo{person}{Feifei Chen}, \bibinfo{person}{Chunyang Chen}, \bibinfo{person}{Faliang Huang}, \bibinfo{person}{Hai Jin}, {and} \bibinfo{person}{Yun Yang}.} \bibinfo{year}{2023}\natexlab{a}.
\newblock \showarticletitle{FlexiFed: Personalized Federated Learning for Edge Clients with Heterogeneous Model Architectures}. In \bibinfo{booktitle}{\emph{Proceedings of the {ACM} Web Conference 2023, {WWW} 2023, Austin, TX, USA, 30 April 2023 - 4 May 2023}}, \bibfield{editor}{\bibinfo{person}{Ying Ding}, \bibinfo{person}{Jie Tang}, \bibinfo{person}{Juan~F. Sequeda}, \bibinfo{person}{Lora Aroyo}, \bibinfo{person}{Carlos Castillo}, {and} \bibinfo{person}{Geert{-}Jan Houben}} (Eds.). \bibinfo{publisher}{{ACM}}, \bibinfo{pages}{2979--2990}.
\newblock
\urldef\tempurl%
\url{https://doi.org/10.1145/3543507.3583347}
\showDOI{\tempurl}


\bibitem[Wang et~al\mbox{.}(2024)]%
        {icdeFa}
\bibfield{author}{\bibinfo{person}{Yong Wang}, \bibinfo{person}{Kaiyu Li}, \bibinfo{person}{Guoliang Li}, \bibinfo{person}{Yunyan Guo}, {and} \bibinfo{person}{Zhuo Wan}.} \bibinfo{year}{2024}\natexlab{}.
\newblock \showarticletitle{Fast, Robust and Interpretable Participant Contribution Estimation for Federated Learning}. ICDE.
\newblock


\bibitem[Wang et~al\mbox{.}(2023b)]%
        {logitFusion}
\bibfield{author}{\bibinfo{person}{Yuwei Wang}, \bibinfo{person}{Runhan Li}, \bibinfo{person}{Hao Tan}, \bibinfo{person}{Xuefeng Jiang}, \bibinfo{person}{Sheng Sun}, \bibinfo{person}{Min Liu}, \bibinfo{person}{Bo Gao}, {and} \bibinfo{person}{Zhiyuan Wu}.} \bibinfo{year}{2023}\natexlab{b}.
\newblock \showarticletitle{Federated skewed label learning with logits fusion}.
\newblock \bibinfo{journal}{\emph{arXiv preprint arXiv:2311.08202}} (\bibinfo{year}{2023}).
\newblock


\bibitem[Wang et~al\mbox{.}(2019)]%
        {symmetricCE}
\bibfield{author}{\bibinfo{person}{Yisen Wang}, \bibinfo{person}{Xingjun Ma}, \bibinfo{person}{Zaiyi Chen}, \bibinfo{person}{Yuan Luo}, \bibinfo{person}{Jinfeng Yi}, {and} \bibinfo{person}{James Bailey}.} \bibinfo{year}{2019}\natexlab{}.
\newblock \showarticletitle{Symmetric Cross Entropy for Robust Learning With Noisy Labels}. In \bibinfo{booktitle}{\emph{2019 {IEEE/CVF} International Conference on Computer Vision, {ICCV} 2019, Seoul, Korea (South), October 27 - November 2, 2019}}. \bibinfo{publisher}{{IEEE}}, \bibinfo{pages}{322--330}.
\newblock
\urldef\tempurl%
\url{https://doi.org/10.1109/ICCV.2019.00041}
\showDOI{\tempurl}


\bibitem[Wei et~al\mbox{.}(2022)]%
        {cifarn}
\bibfield{author}{\bibinfo{person}{Jiaheng Wei}, \bibinfo{person}{Zhaowei Zhu}, \bibinfo{person}{Hao Cheng}, \bibinfo{person}{Tongliang Liu}, \bibinfo{person}{Gang Niu}, {and} \bibinfo{person}{Yang Liu}.} \bibinfo{year}{2022}\natexlab{}.
\newblock \showarticletitle{Learning with Noisy Labels Revisited: {A} Study Using Real-World Human Annotations}. In \bibinfo{booktitle}{\emph{The Tenth International Conference on Learning Representations, {ICLR} 2022, Virtual Event, April 25-29, 2022}}. \bibinfo{publisher}{OpenReview.net}.
\newblock
\urldef\tempurl%
\url{https://openreview.net/forum?id=TBWA6PLJZQm}
\showURL{%
\tempurl}


\bibitem[Wu et~al\mbox{.}(2023)]%
        {FedNoRo}
\bibfield{author}{\bibinfo{person}{Nannan Wu}, \bibinfo{person}{Li Yu}, \bibinfo{person}{Xuefeng Jiang}, \bibinfo{person}{Kwang{-}Ting Cheng}, {and} \bibinfo{person}{Zengqiang Yan}.} \bibinfo{year}{2023}\natexlab{}.
\newblock \showarticletitle{FedNoRo: Towards Noise-Robust Federated Learning by Addressing Class Imbalance and Label Noise Heterogeneity}. In \bibinfo{booktitle}{\emph{Proceedings of the Thirty-Second International Joint Conference on Artificial Intelligence, {IJCAI} 2023, 19th-25th August 2023, Macao, SAR, China}}. \bibinfo{publisher}{ijcai.org}, \bibinfo{pages}{4424--4432}.
\newblock
\urldef\tempurl%
\url{https://doi.org/10.24963/ijcai.2023/492}
\showDOI{\tempurl}


\bibitem[Xiao et~al\mbox{.}(2015)]%
        {clothing1m}
\bibfield{author}{\bibinfo{person}{Tong Xiao}, \bibinfo{person}{Tian Xia}, \bibinfo{person}{Yi Yang}, \bibinfo{person}{Chang Huang}, {and} \bibinfo{person}{Xiaogang Wang}.} \bibinfo{year}{2015}\natexlab{}.
\newblock \showarticletitle{Learning from massive noisy labeled data for image classification}. In \bibinfo{booktitle}{\emph{{IEEE} Conference on Computer Vision and Pattern Recognition, {CVPR} 2015, Boston, MA, USA, June 7-12, 2015}}. \bibinfo{publisher}{{IEEE} Computer Society}, \bibinfo{pages}{2691--2699}.
\newblock
\urldef\tempurl%
\url{https://doi.org/10.1109/CVPR.2015.7298885}
\showDOI{\tempurl}


\bibitem[Xu et~al\mbox{.}(2022)]%
        {fedcorr}
\bibfield{author}{\bibinfo{person}{Jingyi Xu}, \bibinfo{person}{Zihan Chen}, \bibinfo{person}{Tony Q.~S. Quek}, {and} \bibinfo{person}{Kai Fong~Ernest Chong}.} \bibinfo{year}{2022}\natexlab{}.
\newblock \showarticletitle{FedCorr: Multi-Stage Federated Learning for Label Noise Correction}. In \bibinfo{booktitle}{\emph{{IEEE/CVF} Conference on Computer Vision and Pattern Recognition, {CVPR} 2022, New Orleans, LA, USA, June 18-24, 2022}}. \bibinfo{publisher}{{IEEE}}, \bibinfo{pages}{10174--10183}.
\newblock
\urldef\tempurl%
\url{https://doi.org/10.1109/CVPR52688.2022.00994}
\showDOI{\tempurl}


\bibitem[Xue et~al\mbox{.}(2023)]%
        {biad}
\bibfield{author}{\bibinfo{person}{Jingjing Xue}, \bibinfo{person}{Min Liu}, \bibinfo{person}{Sheng Sun}, \bibinfo{person}{Yuwei Wang}, \bibinfo{person}{Hui Jiang}, {and} \bibinfo{person}{Xuefeng Jiang}.} \bibinfo{year}{2023}\natexlab{}.
\newblock \showarticletitle{FedBIAD: Communication-Efficient and Accuracy-Guaranteed Federated Learning with Bayesian Inference-Based Adaptive Dropout}. In \bibinfo{booktitle}{\emph{{IEEE} International Parallel and Distributed Processing Symposium, {IPDPS} 2023, St. Petersburg, FL, USA, May 15-19, 2023}}. \bibinfo{publisher}{{IEEE}}, \bibinfo{pages}{489--500}.
\newblock
\urldef\tempurl%
\url{https://doi.org/10.1109/IPDPS54959.2023.00056}
\showDOI{\tempurl}


\bibitem[Yan et~al\mbox{.}(2024)]%
        {fedeye}
\bibfield{author}{\bibinfo{person}{Bingjie Yan}, \bibinfo{person}{Danmin Cao}, \bibinfo{person}{Xinlong Jiang}, \bibinfo{person}{Yiqiang Chen}, \bibinfo{person}{Weiwei Dai}, \bibinfo{person}{Fan Dong}, \bibinfo{person}{Wuliang Huang}, \bibinfo{person}{Teng Zhang}, \bibinfo{person}{Chenlong Gao}, \bibinfo{person}{Qian Chen}, {et~al\mbox{.}}} \bibinfo{year}{2024}\natexlab{}.
\newblock \showarticletitle{FedEYE: A scalable and flexible end-to-end federated learning platform for ophthalmology}.
\newblock \bibinfo{journal}{\emph{Patterns}} \bibinfo{volume}{5}, \bibinfo{number}{2} (\bibinfo{year}{2024}).
\newblock


\bibitem[Yang et~al\mbox{.}(2020)]%
        {recomm}
\bibfield{author}{\bibinfo{person}{Liu Yang}, \bibinfo{person}{Ben Tan}, \bibinfo{person}{Vincent~W Zheng}, \bibinfo{person}{Kai Chen}, {and} \bibinfo{person}{Qiang Yang}.} \bibinfo{year}{2020}\natexlab{}.
\newblock \showarticletitle{Federated recommendation systems}.
\newblock \bibinfo{journal}{\emph{Federated Learning: Privacy and Incentive}} (\bibinfo{year}{2020}), \bibinfo{pages}{225--239}.
\newblock


\bibitem[Yang et~al\mbox{.}(2022)]%
        {robustfl}
\bibfield{author}{\bibinfo{person}{Seunghan Yang}, \bibinfo{person}{Hyoungseob Park}, \bibinfo{person}{Junyoung Byun}, {and} \bibinfo{person}{Changick Kim}.} \bibinfo{year}{2022}\natexlab{}.
\newblock \showarticletitle{Robust Federated Learning With Noisy Labels}.
\newblock \bibinfo{journal}{\emph{{IEEE} Intell. Syst.}} \bibinfo{volume}{37}, \bibinfo{number}{2} (\bibinfo{year}{2022}), \bibinfo{pages}{35--43}.
\newblock
\urldef\tempurl%
\url{https://doi.org/10.1109/MIS.2022.3151466}
\showDOI{\tempurl}


\bibitem[Yi and Wu(2019)]%
        {pencil}
\bibfield{author}{\bibinfo{person}{Kun Yi} {and} \bibinfo{person}{Jianxin Wu}.} \bibinfo{year}{2019}\natexlab{}.
\newblock \showarticletitle{Probabilistic end-to-end noise correction for learning with noisy labels}. In \bibinfo{booktitle}{\emph{Proceedings of the IEEE/CVF conference on computer vision and pattern recognition}}. \bibinfo{pages}{7017--7025}.
\newblock


\bibitem[Yin et~al\mbox{.}(2018)]%
        {TrimmedMean}
\bibfield{author}{\bibinfo{person}{Dong Yin}, \bibinfo{person}{Yudong Chen}, \bibinfo{person}{Kannan Ramchandran}, {and} \bibinfo{person}{Peter~L. Bartlett}.} \bibinfo{year}{2018}\natexlab{}.
\newblock \showarticletitle{Byzantine-Robust Distributed Learning: Towards Optimal Statistical Rates}. In \bibinfo{booktitle}{\emph{Proceedings of the 35th International Conference on Machine Learning, {ICML} 2018, Stockholmsm{\"{a}}ssan, Stockholm, Sweden, July 10-15, 2018}} \emph{(\bibinfo{series}{Proceedings of Machine Learning Research}, Vol.~\bibinfo{volume}{80})}, \bibfield{editor}{\bibinfo{person}{Jennifer~G. Dy} {and} \bibinfo{person}{Andreas Krause}} (Eds.). \bibinfo{publisher}{{PMLR}}, \bibinfo{pages}{5636--5645}.
\newblock
\urldef\tempurl%
\url{http://proceedings.mlr.press/v80/yin18a.html}
\showURL{%
\tempurl}


\bibitem[Yu et~al\mbox{.}(2019)]%
        {co-teaching+}
\bibfield{author}{\bibinfo{person}{Xingrui Yu}, \bibinfo{person}{Bo Han}, \bibinfo{person}{Jiangchao Yao}, \bibinfo{person}{Gang Niu}, \bibinfo{person}{Ivor~W. Tsang}, {and} \bibinfo{person}{Masashi Sugiyama}.} \bibinfo{year}{2019}\natexlab{}.
\newblock \showarticletitle{How does Disagreement Help Generalization against Label Corruption?}. In \bibinfo{booktitle}{\emph{Proceedings of the 36th International Conference on Machine Learning, {ICML} 2019, 9-15 June 2019, Long Beach, California, {USA}}} \emph{(\bibinfo{series}{Proceedings of Machine Learning Research}, Vol.~\bibinfo{volume}{97})}, \bibfield{editor}{\bibinfo{person}{Kamalika Chaudhuri} {and} \bibinfo{person}{Ruslan Salakhutdinov}} (Eds.). \bibinfo{publisher}{{PMLR}}, \bibinfo{pages}{7164--7173}.
\newblock
\urldef\tempurl%
\url{http://proceedings.mlr.press/v97/yu19b.html}
\showURL{%
\tempurl}


\bibitem[Zhang et~al\mbox{.}(2023)]%
        {fedloke}
\bibfield{author}{\bibinfo{person}{Chao Zhang}, \bibinfo{person}{Fangzhao Wu}, \bibinfo{person}{Jingwei Yi}, \bibinfo{person}{Derong Xu}, \bibinfo{person}{Yang Yu}, \bibinfo{person}{Jindong Wang}, \bibinfo{person}{Yidong Wang}, \bibinfo{person}{Tong Xu}, \bibinfo{person}{Xing Xie}, {and} \bibinfo{person}{Enhong Chen}.} \bibinfo{year}{2023}\natexlab{}.
\newblock \showarticletitle{Non-IID always Bad? Semi-Supervised Heterogeneous Federated Learning with Local Knowledge Enhancement}. In \bibinfo{booktitle}{\emph{Proceedings of the 32nd ACM International Conference on Information and Knowledge Management}}. \bibinfo{pages}{3257--3267}.
\newblock


\bibitem[Zhang et~al\mbox{.}(2018)]%
        {mixup}
\bibfield{author}{\bibinfo{person}{Hongyi Zhang}, \bibinfo{person}{Moustapha Ciss{\'{e}}}, \bibinfo{person}{Yann~N. Dauphin}, {and} \bibinfo{person}{David Lopez{-}Paz}.} \bibinfo{year}{2018}\natexlab{}.
\newblock \showarticletitle{mixup: Beyond Empirical Risk Minimization}. In \bibinfo{booktitle}{\emph{6th International Conference on Learning Representations, {ICLR} 2018, Vancouver, BC, Canada, April 30 - May 3, 2018, Conference Track Proceedings}}. \bibinfo{publisher}{OpenReview.net}.
\newblock
\urldef\tempurl%
\url{https://openreview.net/forum?id=r1Ddp1-Rb}
\showURL{%
\tempurl}


\bibitem[Zhang et~al\mbox{.}(2021)]%
        {incentives}
\bibfield{author}{\bibinfo{person}{Jingwen Zhang}, \bibinfo{person}{Yuezhou Wu}, {and} \bibinfo{person}{Rong Pan}.} \bibinfo{year}{2021}\natexlab{}.
\newblock \showarticletitle{Incentive Mechanism for Horizontal Federated Learning Based on Reputation and Reverse Auction}. In \bibinfo{booktitle}{\emph{{WWW} '21: The Web Conference 2021, Virtual Event / Ljubljana, Slovenia, April 19-23, 2021}}, \bibfield{editor}{\bibinfo{person}{Jure Leskovec}, \bibinfo{person}{Marko Grobelnik}, \bibinfo{person}{Marc Najork}, \bibinfo{person}{Jie Tang}, {and} \bibinfo{person}{Leila Zia}} (Eds.). \bibinfo{publisher}{{ACM} / {IW3C2}}, \bibinfo{pages}{947--956}.
\newblock
\urldef\tempurl%
\url{https://doi.org/10.1145/3442381.3449888}
\showDOI{\tempurl}


\bibitem[Zhao et~al\mbox{.}(2022)]%
        {adaptiveLA}
\bibfield{author}{\bibinfo{person}{Yan Zhao}, \bibinfo{person}{Weicong Chen}, \bibinfo{person}{Xu Tan}, \bibinfo{person}{Kai Huang}, {and} \bibinfo{person}{Jihong Zhu}.} \bibinfo{year}{2022}\natexlab{}.
\newblock \showarticletitle{Adaptive logit adjustment loss for long-tailed visual recognition}. In \bibinfo{booktitle}{\emph{Proceedings of the AAAI conference on artificial intelligence}}, Vol.~\bibinfo{volume}{36}. \bibinfo{pages}{3472--3480}.
\newblock


\bibitem[Zheng et~al\mbox{.}(2023)]%
        {fedpse}
\bibfield{author}{\bibinfo{person}{Longfei Zheng}, \bibinfo{person}{Yingting Liu}, \bibinfo{person}{Xiaolong Xu}, \bibinfo{person}{Chaochao Chen}, \bibinfo{person}{Yuzhou Tang}, \bibinfo{person}{Lei Wang}, {and} \bibinfo{person}{Xiaolong Hu}.} \bibinfo{year}{2023}\natexlab{}.
\newblock \showarticletitle{FedPSE: Personalized Sparsification with Element-wise Aggregation for Federated Learning}. In \bibinfo{booktitle}{\emph{Proceedings of the 32nd ACM International Conference on Information and Knowledge Management}}. \bibinfo{pages}{3514--3523}.
\newblock


\end{thebibliography}

% \appendix






\end{document}